\documentclass[english]{article}
\usepackage[T1]{fontenc}
\usepackage[latin1]{inputenc}
\usepackage{geometry}
\usepackage{float}
\usepackage{mathrsfs}
\usepackage{amsmath}
\usepackage{amssymb}
\usepackage{amsthm}
\usepackage{graphicx}
\usepackage{subfig}
\usepackage{booktabs}
\usepackage{multirow}

\makeatletter

\floatstyle{ruled}
\newfloat{algorithm}{tbp}{loa}
\providecommand{\algorithmname}{Algorithm}
\floatname{algorithm}{\protect\algorithmname}

\usepackage{amsthm}

\usepackage{mathrsfs}

\usepackage{amsfonts}

\usepackage{epsfig}

\usepackage{bm}

\usepackage{mathrsfs}

\usepackage{enumerate}

\@ifundefined{definecolor}{\@ifundefined{definecolor}
 {\@ifundefined{definecolor}
 {\usepackage{color}}{}
}{}
}{}

\usepackage{subfig}
\usepackage[all]{xy}

\newtheorem{rem}{Remark}[section]
\newtheorem{prop}{Proposition}[section]

\newcounter{hypA}

\newcounter{hypB}

\newcounter{hypD}
\newenvironment{hypD}{\refstepcounter{hypD}\begin{itemize}
 \item[({\bf D\arabic{hypD}})]}{\end{itemize}}

\usepackage{babel}\date{}

\usepackage{babel}

\makeatother

\usepackage{babel}

\begin{document}

\begin{center}

{\Large \textbf{Unbiased Estimation of the Gradient of the  Log-Likelihood for a Class of Continuous-Time State-Space Models}}

\vspace{0.5cm}

BY MARCO BALLESIO \& AJAY JASRA

{\footnotesize Computer, Electrical and Mathematical Sciences and Engineering Division, King Abdullah University of Science and Technology, Thuwal, 23955, KSA.}
{\footnotesize E-mail:\,} \texttt{\emph{\footnotesize marco.ballesio@kaust.edu.sa, ajay.jasra@kaust.edu.sa}}
\end{center}

\begin{abstract}
In this paper, we consider static parameter estimation for a class of continuous-time state-space models. Our goal is to obtain an unbiased estimate of the 
gradient of the 
log-likelihood (score function), which is an estimate that is unbiased even if the stochastic processes involved in the model must be discretized in time. To achieve this goal, we apply a \emph{doubly randomized scheme} (see, e.g.,~\cite{ub_mcmc, ub_grad}), that involves a novel coupled conditional particle filter (CCPF) on the second level of randomization \cite{jacob2}. Our novel estimate helps facilitate the application
of gradient-based estimation algorithms, such as stochastic-gradient Langevin descent. We illustrate our methodology in the context of stochastic gradient descent (SGD) in several numerical examples and compare with the Rhee \& Glynn estimator \cite{rhee,vihola}. 
\end{abstract}

\section{Introduction}

State-space models are used in many applications in applied mathematics, statistics, and economics (see, e.g.,~\cite{cappe}). They typically comprise a hidden or unobserved Markov chain that is
associated with an observation process. In many cases of practical interest, there are unknown finite-dimensional parameters, $\theta\in\Theta\subset\mathbb{R}^{d_{\theta}}$, that characterize the dynamics of the hidden and observed processes. The objective of this paper is to consider the estimation of these parameters on the basis of a fixed-length dataset, when the observations and hidden process are both diffusion processes.

There are many challenges in parameter estimation for the class of continuous-time state-space models under consideration. The first challenge is that in practice, data are not observed in continuous time; thus, it is necessary to perform time-discretization (e.g.,~Euler-Maruyama method) of the observation process at the very least. The second challenge is that the hidden diffusion process may often be unavailable (e.g.,~for exact simulation) without also using time discretization. The third challenge is that even under the aforementioned approximations, to compute the log-likelihood function or its gradient with respect to~$\theta$ (the score function), which is the estimation paradigm that is followed in this paper, it is still not possible to compute these quantities analytically. We proceed under the assumption that one must time-discretize both the observation and hidden process and that one seeks the parameters that maximize the log-likelihood function (the result of which is the maximum likelihood estimator (MLE)). We use a particular identity for the score function that is provided in \cite{campillo} and based on the Girsanov change of measure. Alternative identities are discussed in \cite{beskos} but are not considered in this paper.

Given the problem under study, there exist several mechanisms for computing the MLE; however, but we restrict ourselves to gradient-based algorithms, that is, iterative algorithms
that compute estimates of $\theta$ using the score function. Then, the objective is to estimate the score function for any given $\theta$. We remark, however, that to ensure convergence of the gradient algorithm, it is often preferable to produce an unbiased stochastic estimate of the score. It is well known that ensuring the convergence of stochastic gradient methods is simpler when the estimate of the gradient is unbiased (see, e.g.,~\cite{BMP90}).

In the context of state-space models in discrete and continuous time, there already exists substantial literature on score estimation (see,  e.g.,~\cite{beskos,backward,forward_smoothing,poyia}). Most of these techniques are based on sequential Monte Carlo (SMC) algorithms (see \cite{chopin} for an introduction), which are simulation-based methods that use a collection of $N\geq 1$ samples generated in parallel and sequentially in time. 
For the problem of interest, when these algorithms can be applied, they produce consistent estimates of the score function (in terms of
the number of samples $N$), but they will typically introduce
a bias with respect to the time discretization. The aim of this paper is to address this problem. 

Intrinsically, the problem of unbiased estimation of the score function can be placed within the context of exact estimation of the (ratios of) expectations with respect to diffusion processes. 
The topic of unbiased estimation of the expectation associated with  diffusion processes has received considerable attention in recent years. The approaches can be roughly divided into two distinct categories: one that focuses on exact simulation of the diffusion of interest \cite{beskos1,beskos2} (see also \cite{blanchet}), and another that is based on randomization schemes \cite{mcl,rhee}.
The first class of methodologies is based on an elegant paradigm constructing unbiased estimators using the underlying properties of the diffusion process. Due to its nature, however,
this class of methodologies cannot be applied for every diffusion process. The second method is arguably more universally applicable and is the focus in this paper. The approach of 
\cite{mcl,rhee} places a probability distribution over the level of time discretization and is sufficient (but not necessary) to unbiasedly estimate differences of expectations with respect to laws of the time-discretized diffusion process to obtain an unbiased and finite-variance estimator of the expectation with respect to the law of the original diffusion process. 

As mentioned above, in the case of score estimation, there is no  expectation, but rather a ratio of expectations which takes us out of the original context in \cite{mcl,rhee}. The approach that we use in this paper is to follow \cite{ub_mcmc, ub_grad} to consider a so-called \emph{doubly randomized scheme}. The first level of discretization is as in 
\cite{mcl,rhee}: however, the second level of randomization is derived using a new type of coupled conditional particle filter (CCPF) \cite{jacob2} that provides an unbiased estimation
of (differences of) ratios of expectations of the diffusion processes as required. This principle was developed in \cite{ub_grad}: however, it was applied for discrete-time observations, not continuous-time observations.
The main contribution of this paper is to extend the methodology of \cite{ub_grad} to a new class of models and to implement it in several challenging examples.

The remainder of this paper is structured as follows. In Section \ref{sec:problem}, we formalize the problem of interest while in Section \ref{sec:app}, we describe our proposed approach.
In Section \ref{sec:simos}, we present numerical results, which illustrate the utility of our methodology.

\section{Problem}\label{sec:problem}

\subsection{Notations}

Let $(\mathsf{X},\mathcal{X})$ be a measurable space.
For $\varphi:\mathsf{X}\rightarrow\mathbb{R}$, we write $\mathcal{B}_b(\mathsf{X})$ to denote the collection of bounded measurable functions. 
Let $\varphi:\mathbb{R}^d\rightarrow\mathbb{R}$, $\textrm{Lip}_{\|\cdot\|_2}(\mathbb{R}^{d})$ denote the collection of real-valued functions that are Lipschitz with respect to $\|\cdot\|_2$ ($\|\cdot\|_p$ denotes the $\mathbb{L}_p$-norm of a vector $x\in\mathbb{R}^d$). That is, $\varphi\in\textrm{Lip}_{\|\cdot\|_2}(\mathbb{R}^{d})$ if there exists $C<+\infty$ such that for any $(x,y)\in\mathbb{R}^{2d}$
$$
|\varphi(x)-\varphi(y)| \leq C\|x-y\|_2.
$$
For $\varphi\in\mathcal{B}_b(\mathsf{X})$, we write the supremum norm $\|\varphi\|=\sup_{x\in\mathsf{X}}|\varphi(x)|$.
$\mathcal{P}(\mathsf{X})$  denotes the collection of probability measures on $(\mathsf{X},\mathcal{X})$.
For measure $\mu$ on $(\mathsf{X},\mathcal{X})$
and $\varphi\in\mathcal{B}_b(\mathsf{X})$, the notation $\mu(\varphi)=\int_{\mathsf{X}}\varphi(x)\mu(dx)$ is used. 
$B(\mathbb{R}^d)$ denotes the Borel sets on $\mathbb{R}^d$.
Let $K:\mathsf{X}\times\mathcal{X}\rightarrow[0,1]$ be a Markov kernel and $\mu$ be a measure: then, we use the notation
$
\mu K(dy) = \int_{\mathsf{X}}\mu(dx) K(x,dy)
$
and for $\varphi\in\mathcal{B}_b(\mathsf{X})$, 
$
K(\varphi)(x) = \int_{\mathsf{X}} \varphi(y) K(x,dy).
$
For $A\in\mathcal{X}$, the indicator is written as $\mathbb{I}_A(x)$.
$\mathcal{U}_A$ denotes the uniform distribution on set~$A$. 
$\mathcal{N}_s(\mu,\Sigma)$ (resp.~$\psi_s(x;\mu,\Sigma)$)
denotes an $s$-dimensional Gaussian distribution (density evaluated at $x\in\mathbb{R}^s$) of mean $\mu$ and covariance $\Sigma$. If $s=1$, we omit subscript $s$. For a vector/matrix $X$, $X^*$ is used to denote the transpose of $X$.
For $A\in\mathcal{X}$, $\delta_A(du)$ denotes the Dirac measure of $A$, and if $A=\{x\}$ with $x\in \mathsf{X}$, we write $\delta_x(du)$. 
For a vector-valued function in $d$ dimensions (resp.~$d$-dimensional vector), such as $\varphi(x)$ (resp.~$x$), we write the $i$-{\textrm{th}} component ($i\in\{1,\dots,d\}$) as $\varphi^{(i)}(x)$ (resp.~$x^i$). For a $d\times q$ matrix $x$, we write the $(i,j)$-{\textrm{th}} entry as $x^{(ij)}$.

\subsection{Model}

Let $(\Omega,\mathcal{F},\{\mathcal{F}_t\}_{t\geq 0}, \mathbb{P}_{\theta})$ be a filtered probability space. Let $\theta\in\Theta\subset\mathbb{R}^{d_{\theta}}$, with $\Theta$ compact, $d_{\theta}\in\mathbb{N}$ and $d_{\theta}<+\infty$ such that $\{\mathbb{P}_{\theta}:\theta\in\Theta\}$ defines a collection of probability spaces. We consider a pair of stochastic processes $\{Y_t\}_{t\geq 0}$, $\{X_t\}_{t\geq 0}$, with $Y_t\in\mathbb{R}^{d_y}$, $X_t\in\mathbb{R}^{d_x}$ $(d_{y},d_x)\in\mathbb{N}^2$, $d_{x},d_y<+\infty$, with $X_0=x_*\in\mathbb{R}^{d_x}$ given:
\begin{eqnarray}
dY_t & = & h_{\theta}(X_t)dt + dB_t \label{eq:obs}\\
dX_t & = & b_{\theta}(X_t)dt + \sigma(X_t)dW_t \label{eq:state}
\end{eqnarray}
where for each $\theta\in\Theta$, $h_{\theta}:\mathbb{R}^{d_x}\rightarrow\mathbb{R}^{d_y}$, $b_{\theta}:\mathbb{R}^{d_x}\rightarrow\mathbb{R}^{d_x}$, $\sigma:\mathbb{R}^{d_x}\rightarrow\mathbb{R}^{d_x\times d_x}$ with $\sigma$ of full rank, $X_0=x_*$ is given, and $\{B_t\}_{t\geq 0}, \{W_t\}_{t\geq 0}$
are independent standard Brownian motions of dimension $d_y$ and $d_x$, respectively. Note that, we can place a probability on $X_0$ and if we do, we denote it $\mu$ (independent of $\theta$) - for now
we simply take $\mu(dx_0)=\delta_{\{x_*\}}(dx_0)$.

To minimize any technical difficulties, the following assumption is made throughout this paper:
\begin{hypD}\label{hyp_diff:1}
We have the following:
\begin{enumerate}
\item{$\sigma$ is continuous and bounded, and $a(x):=\sigma(x)\sigma(x)^*$ is uniformly elliptic.}
\item{For each $\theta\in\Theta$, $(h_\theta,b_{\theta})$ are bounded, measurable, and $h_{\theta}^{(i)}\in\textrm{Lip}_{\|\cdot\|_2}(\mathbb{R}^{d_x})$, $i\in\{1,\dots,d_y\}$.}
\item{$h_{\theta},b_{\theta}$ are continuously differentiable with respect to $\theta$, and for each $\theta\in\Theta$, $\nabla_{\theta}h_\theta:\mathbb{R}^{d_x}\rightarrow\mathbb{R}^{d_y\times d_{\theta}}$, $\nabla_{\theta}b_\theta:\mathbb{R}^{d_x}\rightarrow\mathbb{R}^{d_x\times d_{\theta}}$, with
$(\nabla_{\theta}h_\theta,\nabla_{\theta}b_{\theta})$ bounded and measurable, and $\nabla_{\theta}h_{\theta}^{(ij)}\in\textrm{Lip}_{\|\cdot\|_2}(\mathbb{R}^{d_x})$, $(i,j)\in\{1,\dots,d_y\}\times\{1,\dots,d_{\theta}\}$.}
\item{$\phi_{\theta}(x):=[\nabla_{\theta}b_{\theta}]^*(x)a(x)^{-1}\sigma(x)$. For each $\theta\in\Theta$, $\phi_{\theta}^{(ij)}\in\textrm{Lip}_{\|\cdot\|_2}(\mathbb{R}^{d_x})$, $(i,j)\in\{1,\dots,d_{\theta}\}\times\{1,\dots,d_{x}\}$.}
\end{enumerate}
\end{hypD}

Now, we introduce the probability measure $\overline{\mathbb{P}}_{\theta}$, which is equivalent to $\mathbb{P}_{\theta}$ defined by the Radon-Nikodym derivative 
$$
Z_{t,\theta}:=\frac{d\mathbb{P}_{\theta}}{d\overline{\mathbb{P}}_{\theta}}\bigg\rvert_{\mathcal{F}_{t}} = \exp\Big\{\int_{0}^t h_{\theta}(X_s)^*dY_s - \frac{1}{2}\int_{0}^t h_{\theta}(X_s)^*h_{\theta}(X_s)ds\Big\}
$$
with $\{X_t\}_{t\geq 0}$ following the dynamics \eqref{eq:state} and $\{Y_t\}_{t\geq 0}$ solving the dynamics $dY_{t}=dB_t$ under $\overline{\mathbb{P}}_{\theta}$. 
The Girsanov theorem states that for any $\varphi\in\mathcal{B}_b(\mathbb{R}^{d_x})$ that satisfies (D\ref{hyp_diff:1}), it holds that
\begin{equation}
\mathbb{E}_{\theta}\Big[\varphi(X_t)|\mathcal{Y}_t\Big]=\overline{\mathbb{E}}_{\theta}\Big[\varphi(X_t)Z_{t,\theta}|\mathcal{Y}_t\Big],
\end{equation}
where $\mathcal{Y}_t$ is the filtration generated by the process $\{Y_s\}_{0\leq s\leq t}$. We define the solution of the Zakai equation for $\varphi\in\mathcal{B}_b(\mathbb{R}^{d_x})$ as
$$
\gamma_{t,\theta}(\varphi) := \overline{\mathbb{E}}_{\theta}\Big[\varphi(X_t)Z_{t,\theta}|\mathcal{Y}_t\Big].
$$
Our objective is to, almost surely, unbiasedly estimate the gradient of the log-likelihood $\nabla_{\theta}\log(\gamma_{T,\theta}(1))$. Adding minor regularity conditions on coefficients (see, e.g.,~\cite{campillo}),
\begin{equation}\label{eq:gll}
\nabla_{\theta}\log(\gamma_{T,\theta}(1)) = \frac{\overline{\mathbb{E}}_{\theta}[\lambda_{T,\theta}Z_{T,\theta}|\mathcal{Y}_{T}]}{\overline{\mathbb{E}}_{\theta}[Z_{T,\theta}|\mathcal{Y}_{T}]} 
\end{equation}
where
$$
\lambda_{T,\theta} = \int_{0}^T[\nabla_{\theta}b_{\theta}(X_s)]^*a(X_s)^{-1}\sigma(X_s)dW_s + \int_{0}^T[\nabla_{\theta}h_{\theta}(X_s)]^* dY_s - \int_{0}^T[\nabla_{\theta}h_{\theta}(X_s)]^*h_{\theta}(X_s)ds.
$$
We assume throughout this paper that $T\in\mathbb{N}$. Note that \cite{beskos} derives an alternative expression to \eqref{eq:gll} that does not require $\sigma$ to be independent of
$\theta$; however, its approximation is significantly more complex than we consider.


\subsection{Discretized Model}

In practice, we must work with a discretization of the model in \eqref{eq:obs}-\eqref{eq:state} since an analytic solution of \eqref{eq:gll} is typically unavailable.
This is because we do not observe data in continuous time and often the exact methods in \cite{beskos1,beskos2}, for example, cannot be applied.
We assume access to the path of data $\{Y_t\}_{0\leq t \leq T}$ up to an (almost) arbitrary level of time discretization. In practice, this would be a very finely observed path, as the former assumption is not possible.

The exposition below closely follows the presentation in \cite{hf_ml}.
Let $l\geq 0$ be given, and consider an Euler discretization of step-size $\Delta_l=2^{-l}$, $k\in\{1,2,\dots,2^lT\}$, $\widetilde{X}_{0}=x_*$:
\begin{eqnarray}
\widetilde{X}_{k\Delta_l} & = & \widetilde{X}_{(k-1)\Delta_l} + b_{\theta}(\widetilde{X}_{(k-1)\Delta_l})\Delta_l + \sigma(\widetilde{X}_{(k-1)\Delta_l})[W_{k\Delta_l}-W_{(k-1)\Delta_l}].\label{eq:disc_state}
\end{eqnarray}
It should be noted that the Brownian motion in \eqref{eq:disc_state} is the same as in \eqref{eq:state} under both $\mathbb{P}_{\theta}$ and $\overline{\mathbb{P}}_{\theta}$.
Set
\begin{eqnarray*}
\lambda_{T,\theta}^l(x_0,x_{\Delta_l},\dots,x_T) & := & \sum_{k=0}^{2^lT-1}\Big\{[\nabla_{\theta}b_{\theta}(x_{k\Delta_l})]^*a(x_{k\Delta_l})^{-1}\sigma(x_{k\Delta_l})[W_{(k+1)\Delta_l}-W_{k\Delta_l}] + \\ & & [\nabla_{\theta}h_{\theta}(x_{k\Delta_l})]^*[Y_{(k+1)\Delta_l}-Y_{k\Delta_l}]
-  [\nabla_{\theta} h_{\theta}(x_{k\Delta_l})]^*h_{\theta}(x_{k\Delta_l})\Delta_l
\Big\}.
\end{eqnarray*}
We remark that when considering \eqref{eq:disc_state}, $\lambda_{T,\theta}^l$ is a function of $(y_0,y_{\Delta_l},\dots,y_T)$ (the dependence on the data is omitted from the notation throughout this paper) and $(\widetilde{X}_0,\widetilde{X}_{\Delta_l},\dots,\widetilde{X}_T)$, as it holds that
\begin{eqnarray*}
[W_{k\Delta_l}-W_{(k-1)\Delta_l}]  & = & \sigma(\widetilde{X}_{(k-1)\Delta_l})^{-1}\Big(\widetilde{X}_{k\Delta_l}- [\widetilde{X}_{(k-1)\Delta_l} + b_{\theta}(\widetilde{X}_{(k-1)\Delta_l})\Delta_l] \Big).
\end{eqnarray*}
Then, for $k\in\{0,1,\dots,2^lT-1\}$, we define
\begin{equation}
G_{k,\theta}^l(x_{k\Delta_l}) := \exp\Big\{h_{\theta}(x_{k\Delta_l})^*(y_{(k+1)\Delta_l}-y_{k\Delta_l})-\frac{\Delta_l}{2}h_{\theta}(x_{k\Delta_l})^*h_{\theta}(x_{k\Delta_l})\Big\}   \label{eq:discret_radon_nikodym}
\end{equation}
and note that
\begin{eqnarray}
Z_{T,\theta}^l(x_0,x_{\Delta_l},\dots,x_{T-\Delta_{l}}) &:=& \prod_{k=0}^{2^lT-1}G_{k,\theta}^l(x_{k\Delta_l}) \nonumber \\
&=& \exp\Big\{\sum_{k=0}^{2^lT-1}\Big[h_{\theta}(x_{k\Delta_l})^*(y_{(k+1)\Delta_l}-y_{k\Delta_l})-\frac{\Delta_l}{2}h_{\theta}(x_{k\Delta_l})^*h_{\theta}(x_{k\Delta_l})\Big]\Big\} \nonumber
\end{eqnarray}
is simply a discretization of $Z_{T,\theta}$. 
We have the discretized approximation of $\nabla_{\theta}\log(\gamma_{T,\theta}(1))$ as follows:
$$
\nabla_{\theta}\log(\gamma_{T,\theta}^l(1)) := \frac{\overline{\mathbb{E}}_{\theta}[\lambda_{T,\theta}^l(\widetilde{X}_0,\widetilde{X}_{\Delta_l},\dots,\widetilde{X}_T)
Z_{T,\theta}^l(\widetilde{X}_0,\widetilde{X}_{\Delta_l},\dots,\widetilde{X}_T)|\mathcal{Y}_T]}{\overline{\mathbb{E}}_{\theta}[Z_{T,\theta}^l(\widetilde{X}_0,\widetilde{X}_{\Delta_l},\dots,\widetilde{X}_T)|\mathcal{Y}_T]}.
$$
The following result, which establishes the convergence of our Euler approximation, is proved in \cite{beskos}. Note that the rate should be $\mathcal{O}(\Delta_l)$, however, this
is not important in the subsequent development of this paper.

\begin{prop}\label{prop:bias}
Assume (D1-3) in \cite{beskos}. Then, for any $(T,\theta)\in[0,\infty)\times\Theta$, there exists $C<+\infty$ such that for any $l\geq 0$, we have
{\footnotesize
$$
\Bigg|\mathbb{E}_{\theta}\Bigg[
\frac{\overline{\mathbb{E}}_{\theta}[\lambda_{T,\theta}Z_{T,\theta}|\mathcal{Y}_T]}{\overline{\mathbb{E}}_{\theta}[Z_{T,\theta}|\mathcal{Y}_T]}
-
\frac{\overline{\mathbb{E}}_{\theta}[\lambda_{T,\theta}^l(\widetilde{X}_0,\widetilde{X}_{\Delta_l},\dots,\widetilde{X}_T)
Z_{T,\theta}^l(\widetilde{X}_0,\widetilde{X}_{\Delta_l},\dots,\widetilde{X}_T)|\mathcal{Y}_T]}{\overline{\mathbb{E}}_{\theta}[Z_{T,\theta}^l(\widetilde{X}_0,\widetilde{X}_{\Delta_l},\dots,\widetilde{X}_T)|\mathcal{Y}_T]}
\Bigg]
\Bigg| \leq C\Delta_l^{1/2}.
$$
}
\end{prop}

\begin{rem}\label{rem:version_prop_rem}
We note that for any real-valued, bounded, and continuous function on the trajectory $\{Y_t\}_{0\leq t \leq T}$, $\varphi$, one can establish, using the proof of Proposition \ref{prop:bias}, that
{\footnotesize
$$
\lim_{l\rightarrow\infty}\Bigg|\mathbb{E}_{\theta}\Bigg[\varphi\Big(\{Y_t\}_{0\leq t \leq T}\Big)\Bigg\{
\frac{\overline{\mathbb{E}}_{\theta}[\lambda_{T,\theta}Z_{T,\theta}|\mathcal{Y}_T]}{\overline{\mathbb{E}}_{\theta}[Z_{T,\theta}|\mathcal{Y}_T]}
-
\frac{\overline{\mathbb{E}}_{\theta}[\lambda_{T,\theta}^l(\widetilde{X}_0,\widetilde{X}_{\Delta_l},\dots,\widetilde{X}_T)
Z_{T,\theta}^l(\widetilde{X}_0,\widetilde{X}_{\Delta_l},\dots,\widetilde{X}_T)|\mathcal{Y}_T]}{\overline{\mathbb{E}}_{\theta}[Z_{T,\theta}^l(\widetilde{X}_0,\widetilde{X}_{\Delta_l},\dots,\widetilde{X}_T)|\mathcal{Y}_T]}\Bigg\}
\Bigg]
\Bigg| = 0.
$$}
\end{rem}

\subsection{Smoothing Identity}

For notational convenience, we drop the $\widetilde{\cdot}$ notation from the Euler discretization, when referring to the subsequent smoothing construction.
$\nabla_{\theta}\log(\gamma_{T,\theta}^l(1))$ can be rewritten as the expectation of $\lambda_{T,\theta}^l(X_0,X_{\Delta_l},\dots,X_T)$ with respect to the smoothing distribution of a discrete-time state-space model. 

Define the probability measure on $(\mathbb{R}^{d_x2^lT},B(\mathbb{R}^{d_x2^lT}))$, recalling that $x_0=x_*$:
\begin{equation}\label{eq:smoother_disc}
\pi_{\theta}^l\big(d(x_{\Delta_l},\dots,x_{T})) := \frac{\big(\prod_{k=0}^{2^lT-1}G_{k,\theta}^l(x_{k\Delta_l})\big)\prod_{k=1}^{2^lT}Q_{\theta}^l(x_{(k-1)\Delta_l},dx_{k\Delta_l})}{\int_{\mathbb{R}^{d_x2^lT}} \big(\prod_{k=0}^{2^lT-1}G_{k,\theta}^l(x_{k\Delta_l})\big)\prod_{k=1}^{2^lT}Q_{\theta}^l(x_{(k-1)\Delta_l},dx_{k\Delta_l})},
\end{equation}
where $Q_{\theta}^l$ is the transition kernel induced from \eqref{eq:disc_state}, and the dependence on the data is suppressed in the notation. Now writing the expectations with respect to $\pi_{\theta}^l$ as $\mathbb{E}_{\pi_{\theta}^l}$, we have
$$
\nabla_{\theta}\log(\gamma_{T,\theta}^l(1)) = \mathbb{E}_{\pi_{\theta}^l}[\lambda_{T,\theta}^l(x_*,X_{\Delta_l},\dots,X_T)].
$$
It is this latter expectation that we use throughout this paper.

\section{Approach}\label{sec:app}

\subsection{Debiasing Schemes}\label{sec:main_idea}

Our objective is to compute an almost surely unbiased estimate of $\nabla_{\theta}\log(\gamma_{T,\theta}(1))$ by only considering $\nabla_{\theta}\log(\gamma_{T,\theta}^l(1))$.
Our construction considers an enlarged probability space $(\Omega^{\star},\mathcal{F}^{\star},\mathbb{P}_{\theta}^{\star})$ associated with $(\Omega,\mathcal{F},\mathbb{P}_{\theta})$ such that the following holds:
\begin{enumerate}
\item{$\mathbb{E}^{\star}_{\theta}[\nabla_{\theta}\log(\gamma_{T,\theta}(1))]=\mathbb{E}_{\theta}[\nabla_{\theta}\log(\gamma_{T,\theta}(1))]$, almost surely
and $\mathbb{E}^{\star}_{\theta}[\nabla_{\theta}\log(\gamma_{T,\theta}^l(1))]=\mathbb{E}_{\theta}[\nabla_{\theta}\log(\gamma_{T,\theta}^l(1))]$.}
\item{One can compute independent random variables $\Psi_{T,\theta}^0,\Psi_{T,\theta}^1,\dots$ such that, almost surely
\begin{equation}
\mathbb{E}^\star_{\theta}[\Psi_{T,\theta}^l] =  \mathbb{E}_{\pi_{\theta}^l}[\lambda_{T,\theta}^l(x_*,X_{\Delta_l},\dots,X_T)] - \mathbb{E}_{\pi_{\theta}^{l-1}}[\lambda_{T,\theta}^{l-1}(x_*,X_{\Delta_{l-1}},\dots,X_T)] \quad l\geq 0  \label{eq:unb_inc}
\end{equation}
with $\mathbb{E}_{\pi_{\theta}^{-1}}[\lambda_{T,\theta}^l(x_*,X_{\Delta_{l-1}},\dots,X_T)]=0$.}

\item{Let $l\in\mathbb{Z}^+$ be a random variable on $(\Omega^{\star},\mathcal{F}^{\star})$ with probability mass function $p^{\star}$, where it is assumed that $p^{\star}(l)>0$ for each $l\geq 0$. Then, we have
$$
\sum_{l=0}^{\infty} \frac{1}{p^{\star}(l)}\mathbb{E}^\star_{\theta}[\|\Psi_{T,\theta}^l\|_2^2] <+\infty.
$$
}
\end{enumerate}

By Proposition \ref{prop:bias}, we know that
$$
\lim_{l\rightarrow\infty}\mathbb{E}^\star_{\theta}[\nabla_{\theta}\log(\gamma_{T,\theta}^l(1))] = \mathbb{E}_{\theta}[\nabla_{\theta}\log(\gamma_{T,\theta}(1))].
$$
Then, one has the following result (see \cite[Theorem 1]{rhee}; see also \cite[Theorem 3]{vihola}):
$$
\mathbb{E}^{\star}_{\theta}\Bigg[\frac{\Psi_{T,\theta}^l}{p^{\star}(l)}\Bigg] = \mathbb{E}^\star_{\theta}[\nabla_{\theta}\log(\gamma_{T,\theta}(1))].
$$
More generally, using this approach, one can deduce that for any real-valued, bounded and continuous function on the trajectory $\{Y_t\}_{0\leq t \leq T}$, $\varphi$ one has
$$
\mathbb{E}_{\theta}^{\star}\Bigg[\frac{\Psi_{T,\theta}^L}{p^{\star}(L)}\varphi\Big(\{Y_t\}_{0\leq t \leq T}\Big)\Bigg] = \mathbb{E}_{\theta}^*\Big[\nabla_{\theta}\log(\gamma_{T,\theta}(1))\varphi\Big(\{Y_t\}_{0\leq t \leq T}\Big)\Big].
$$
That is, 
$$
\mathbb{E}_{\theta}^{\star}\Bigg[\frac{\Psi_{T,\theta}^L}{p^{\star}(L)}\Big|\mathcal{Y}_T\Bigg]
$$
is a version of $\mathbb{E}_{\theta}^*[\nabla_{\theta}\log(\gamma_{T,\theta}(1))|\mathcal{Y}_T]$, i.e.~it is an almost surely unbiased estimator of $\mathbb{E}_{\theta}[\nabla_{\theta}\log(\gamma_{T,\theta}(1))|\mathcal{Y}_T]$. As a result, our
objective is to obtain the random variables $\Psi_{T,\theta}^0,\Psi_{T,\theta}^1,\dots$ so that one can calculate $\Psi_{T,\theta}^L/p^{\star}(L)$. This is the topic of the remainder of the section. In this article, we stress that we have not proved the properties 1.-3.~above, but, it is possible using the analysis in \cite{ub_grad}; we leave the rather substantial proof to future work, but some further discussion is given in Section \ref{sec:theory}. Note also that we consider the so-called single term estimator here, but that can be generalized to the independent term estimator also.


\begin{rem}
To actually compute $\Psi_{T,\theta}^l/p^{\star}(l)$, one expects to have access to a data trajectory that is arbitrarily finely observed (in time), as \eqref{eq:unb_inc} must be satisfied. Typically, this is not possible
in practice; however, we remark that (as we will see) computing $\Psi_{T,\theta}^l/p^{\star}(l)$ is often only possible for $l\leq 50$ due to the computational cost. This drawback is common to all debiasing schemes
(as described in \cite{rhee,vihola}) and thus, we only require very high frequency observations, not an entire trajectory.
\end{rem}

\subsection{Conditional Particle Filter}\label{sec:cpf}

The conditional particle filter is a particle filter that runs conditional on a trajectory $x_{[\Delta_l:T]}\in\mathsf{X}^{T}_l$ with $\mathsf{X}^{T}_l=\mathbb{R}^{d_x 2^l}$ and $x_0=x_\star$.

Setting $u_{[k:k+1]}^{l,i}\in\mathsf{X}_l$ for $i\in\{1,\dots,N\}$, 
\begin{eqnarray*}
F_{k,\theta}^l(i,u_{[k:k+1]}^{l,1:N}) & := &  
\frac{\prod_{m=0}^{\Delta_l^{-1}-1} G_{k+m\Delta_l,\theta}(u_{k+m\Delta_l}^{l,i})}{\sum_{s=1}^N
\prod_{m=0}^{\Delta_l^{-1}-1} G_{k+m\Delta_l,\theta}(u_{k+m\Delta_l}^{l,s})}
\end{eqnarray*}
and $\mathbf{u}_k^{l,i}\in\mathsf{X}_l^{k+1}$, $\mathbf{u}_k^{l,i}=(\mathbf{u}_{k,0}^{l,i},\dots,\mathbf{u}_{k,k}^{l,i})$, $i\in\{1,\dots,N\}$, $k\in\{0,1,\dots,T-1\}$, $\mathbf{u}_{k,j}^{l,i}\in\mathsf{X}_l$, $j\in\{0,1,\dots,k\}$.
We define the CPF kernel $\overline{K}^{l}_{T,\theta}:\mathsf{X}^T_l \rightarrow \mathcal{P}(\mathsf{X}^T_l)$ as
\begin{eqnarray}
\overline{K}^{l}_{T,\theta}\Big( x_{[\Delta_l:T]}, dx'_{[\Delta_l:T]} \Big)=\int_{\mathsf{X}^{NT}_{l}} \overline{\mathbb{K}}^{l}_{T,\theta}\Big(x_{[\Delta_{l}:T]},d\big(\pmb{u}^{l,1:N}_{0},\dots, \mathbf{u}^{l,1:N}_{T-1}\big)\Big)  \nonumber \\
\sum_{s\in\{1,\dots,N\}} F^{l}_{T-1,\theta}\big(s,\mathbf{u}^{l,1:N}_{T-1,T-1}\big)\delta_{\{\mathbf{u}^{l,s}_{T-1}\}}\big(dx'_{[\Delta_l:T]}\big)
\end{eqnarray}
with
$$
\mathbb{\overline{K}}^l_{T,\theta}\Big(x_{[\Delta_l:T]},d(\mathbf{u}_0^{l,1:N},\dots,\mathbf{u}_{T-1}^{l,1:N})\Big) = 
$$
$$
\prod_{i=1}^{N-1}\mathbf{\overline{\overline{\mathbf{Q}}}}_{\theta}^l\big(x,d\mathbf{u}_{0}^{l,i}\big)\delta_{\{x_{[\Delta_l:1]}\}}\big(d\mathbf{u}_0^{l,N}\big)
$$
$$
 \Big[\prod_{k=1}^{T-1} 
\Big\{
\prod_{i=1}^{N-1}\sum_{s\in\{1,\dots,N\}}
F_{k-1,\theta}^{l}(s,\mathbf{u}_{k-1,k-1}^{l,1:N})\times
\mathbf{\overline{Q}}_{k,\theta}^l\big(\mathbf{u}_{k-1}^{l,s},d\mathbf{u}_{k}^{l,i}\big)
\}\delta_{\{x_{[\Delta_l:k]}\}}\Big\}\Big(d\mathbf{u}_{k}^{l,N}\Big)\Big] \label{eq:p_def_ccpf}
$$
probability kernel $\mathbf{\overline{Q}}_{k,\theta}^l:\mathsf{X}^{k}_l\rightarrow \mathcal{P}(\mathsf{X}^{k+1}_{l})$, $u_{[\Delta_l:k]}\in\mathsf{X}^k_l$
\begin{eqnarray*}
\mathbf{\overline{Q}}_{k,\theta}^l\big(u_{[\Delta_l:k]},du_{[\Delta_l:k+1]}'\big) & := &\delta_{\{u_{[\Delta_l:k]}\}}\big(du_{[\Delta_l:k]}'\big)\times  \mathbf{\overline{\overline{Q}}}_{\theta}^l\Big(u_{k}',du_{[k+\Delta_l:k+1]}'\Big)
\end{eqnarray*}
where
\begin{equation}\label{eq:chql_def}
\mathbf{\overline{\overline{Q}}}_{\theta}^l\Big(u_{k}',du_{[k+\Delta_l:k+1]}'\Big)= \prod_{m=1}^{\Delta_{l}^{-1}}Q_{\theta}^l\Big(u_{k+(m-1)\Delta_{l}^{-1}}',du_{k+m\Delta_{l}^{-1}}'\Big).
\end{equation}
The simulation of CPF kernel $\overline{K}^{l}_{T,\theta}:\mathsf{X}^T_l \rightarrow \mathcal{P}(\mathsf{X}^T_l)$ is described in Algorithm \ref{algo:cpf}.

\begin{algorithm}[H]
\begin{enumerate}
\item{Initialize: For $i\in\{1,\dots,N-1\}$ sample $\mathbf{u}_0^{l,i}$ independently using $\mathbf{\overline{\overline{Q}}}_{\theta}^l\big(x,\cdot\big)$. Set $\mathbf{u}_0^{l,N}=x_{[\Delta_l:1]}$, $k=0$.}
\item{Coupled Resampling:  For $i\in\{1,\dots,N-1\}$ sample $r_k^{i}$ accordingly to $F^{l}_{k,\theta}(i,\mathbf{u}^{l,1:N}_{k,k})$.}
\item{Coupled Sampling:  Set $k=k+1$. For $i\in\{1,\dots,N-1\}$ sample $\mathbf{u}_k^{l,i}| \mathbf{u}_{k-1}^{l,r_{k-1}^i}$ conditionally independently using $\mathbf{\overline{Q}}_{k,\theta}^l\Big(\mathbf{u}_{k-1}^{l,r_{k-1}^i},\cdot\Big)$. 
Set $\mathbf{u}_k^{l,N}=x_{[\Delta_l:k+1]}$. If $k=T-1$ go to 4., otherwise go to 2..
}
\item{Select Trajectories: Sample $r^l$ according to $F^{l}_{k,\theta}(i,\mathbf{u}_{T-1,T-1}^{l,1:N})$ (as described, for one
$i$ in 2.).
Return $\mathbf{u}_{T-1}^{l,r^l}$.
}
\end{enumerate}
\caption{Simulating the CPF kernel.}
\label{algo:cpf}
\end{algorithm}

\subsection{Coupled Conditional Particle Filter}\label{sec:ccpf}

We consider the CCPF in \cite{jacob2} (see also \cite{lee} for extensions) that allows one to compute unbiased estimates of expectations with respect to the probability \eqref{eq:smoother_disc}. That is, letting $l\geq 0$, $\theta\in\Theta$ fixed, and $\varphi_{\theta}^l:\mathbb{R}^{d_x2^lT}\rightarrow\mathbb{R}$, $\varphi_{\theta}^l$ being
$\pi_{\theta}^l$-integrable and measurable, 
the CCPF produces an estimate of $\pi_{\theta}^l(\varphi_{\theta}^l)$ that is equal to $\pi_{\theta}^l(\varphi_{\theta}^l)$ in expectation.

Throughout this review of the CCPF, $l\geq 0$ is fixed but finite and $\theta\in\Theta$ is also fixed. Given 
$x_p^l \in\mathbb{R}^{d_x}, x_{p+\Delta_l}^l \in\mathbb{R}^{d_x}, x_q^l \in\mathbb{R}^{d_x}$ , $p\leq q$, $p/\Delta_l\in\mathbb{Z}^+$, and $q/\Delta_l\in\mathbb{Z}^+$, we use the notation $x^l_{[p:q]}:=(x^l_{p},x^l_{p+\Delta_l},\dots,x_q^l)$. The time increment $\Delta_l$ in the subscript  is derived from the superscript $l$ of the vector; when there is no possible confusion, this superscript is omitted from the notation.

\subsubsection{Probability Kernel Coupling} \label{sec:ccpf_kernel_distribution}

To describe the CCPF, we introduce the following coupling of $Q_{\theta}^l$. Suppose that we are given $(x,\mathring{x})\in\mathbb{R}^{d_x}\times\mathbb{R}^{d_x}$; then, $\check{Q}_{\theta}^l:\mathbb{R}^{d_x}\times\mathbb{R}^{d_x}\rightarrow\mathcal{P}(\mathbb{R}^{d_x}\times\mathbb{R}^{d_x})$ is a Markov kernel that is simulated as follows:
\begin{itemize}
\item{Generate $W\sim\mathcal{N}_{d_x}(0,\Delta_l I_{d_x})$.}
\item{Then
\begin{eqnarray*}
X' & = &  x + b_{\theta}(x)\Delta_l + \sigma(x)W \\
\mathring{X}' & = &  \mathring{x} + b_{\theta}(\mathring{x})\Delta_l + \sigma(\mathring{x})W 
\end{eqnarray*}
}
\end{itemize}
$\check{Q}_{\theta}^l\Big((x,\mathring{x}),d(x',\mathring{x}')\Big)$ is such that for any  $(x,\mathring{x})\in\mathbb{R}^{d_x}\times\mathbb{R}^{d_x}$, $A\in B(\mathbb{R}^{d_x})$
$$
\check{Q}_{\theta}^l(A\times \mathbb{R}^{d_x})(x,\mathring{x}) = \int_{A}Q_{\theta}^l(x,dx') \quad \check{Q}_{\theta}^l(\mathbb{R}^{d_x}\times A)(x,\mathring{x}) = \int_{A}Q_{\theta}^l(\mathring{x},d\mathring{x}').
$$ 
We remark that, clearly, if $x=\mathring{x}$, then $x'=\mathring{x}'$.

\subsubsection{CCPF Kernel} \label{sec:ccpf_kernel}

We now introduce the CCPF. Although the CCPF is a particular case of the approach in \cite{jacob2}, the coupled resampling method (also used in \cite{mlpf_new}) can perform
very well in theory \cite{jasra_yu}. The basic principle is to generate a coupled particle filter (see, e.g.,~\cite{jasra_yu,mlpf}) that
runs conditionally on a pair of trajectories $(x_{[\Delta_l:T]},\mathring{x}_{[\Delta_l:T]})\in\mathsf{X}_l^{T}\times\mathsf{X}_l^{T}$, $\mathsf{X}_l:=\mathbb{R}^{d_x2^l}$, since $(x_{0},\mathring{x}_{0})=(x_{\star},x_{\star})$. 
We first introduce an underlying probability kernel $\mathbb{K}^l_{T,\theta}:\mathsf{X}_l^{T}\times\mathsf{X}_l^{T}\rightarrow\mathcal{P}(\mathsf{X}_l^{2NT})$, which is critical in defining the CCPF kernel. Let $(u_{[k:k+1]}^{l,1:N},\mathring{u}_{[k:k+1]}^{l,1:N})\in\mathsf{X}_l^{2N}$, $k\in\{0,1,\dots,T-1\}$, and for $(i,j)\in\{1,\dots,N\}^2$,
{\footnotesize
$$
\omega_{k,\theta}^{l}(i,j,u_{[k:k+1]}^{l,1:N},\mathring{u}_{[k:k+1]}^{l,1:N})  =  
$$
$$
\Big(\sum_{s=1}^N \{F_{k,\theta}^l(s,u_{[k:k+1]}^{l,1:N})\wedge
F_{k,\theta}(s,\mathring{u}_{[k:k+1]}^{l,1:N})\}\Big)
\Bigg(\frac{F_{k,\theta}^l(i,u_{[k:k+1]}^{l,1:N})\wedge
F_{k,\theta}^l(i,\mathring{u}_{[k:k+1]}^{l,1:N})}{\sum_{s=1}^N\{ F_{k,\theta}^l(s,u_{[k:k+1]}^{l,1:N})\wedge
F_{k,\theta}^l(s,\mathring{u}_{[k:k+1]}^{l,1:N})\}}\Bigg)\mathbb{I}_{\{i\}}(j) + 
$$
$$
 \Big(1-\sum_{s=1}^N \{F_{k,\theta}^l(s,u_{[k:k+1]}^{l,1:N})\wedge
F_{k,\theta}^l(s,\mathring{u}_{[k:k+1]}^{l,1:N})\}\Big)
\Bigg(\frac{F_{k,\theta}^l(i,u_{[k:k+1]}^{l,1:N})-F_{k,\theta}^l(i,u_{[k:k+1]}^{l,1:N})\wedge F_{k,\theta}^l(i,\mathring{u}_{[k:k+1]}^{l,1:N})}{1-\sum_{s=1}^N\{F_{k,\theta}^l(s,u_{[k:k+1]}^{l,1:N})\wedge F_{k,\theta}^l(s,\mathring{u}_{[k:k+1]}^{l,1:N})\}}\Bigg)\times 
$$
$$
\Bigg(\frac{F_{k,\theta}^l(j,\mathring{u}_{[k:k+1]}^{l,1:N})-F_{k,\theta}^l(j,u_{[k:k+1]}^{l,1:N})\wedge F_{k,\theta}^l(j,\mathring{u}_{[k:k+1]}^{l,1:N})}{1-\sum_{s=1}^N\{F_{k,\theta}^l(s,u_{[k:k+1]}^{l,1:N})\wedge F_{k,\theta}^l(s,\mathring{u}_{[k:k+1]}^{l,1:N})\}}\Bigg).
$$}
The probability $\omega_{k,\theta}^{l}(i,j,u_{[k:k+1]}^{l,1:N},\mathring{u}_{[k:k+1]}^{l,1:N})$ is simply a maximal
coupling of the resampling probabilities for particular particle filters (see, e.g.,~\cite{mlpf}), which can be performed at $\mathcal{O}(N)$ cost.

Now define the probability kernel, for $k\in\{1,\dots,T-1\}$, $\mathbf{Q}_{k,\theta}^l:\mathsf{X}_l^{2k}\rightarrow\mathcal{P}(\mathsf{X}_l^{2(k+1)})$, $(u_{[\Delta_l:k]},\mathring{u}_{[\Delta_l:k]})\in\mathsf{X}_l^{2k}$ 
\begin{eqnarray*}
\mathbf{Q}_{k,\theta}^l\Big((u_{[\Delta_l:k]},\mathring{u}_{[\Delta_l:k]}),d(u_{[\Delta_l:k+1]}',\mathring{u}_{[\Delta_l:k+1]}')\Big) & := &\delta_{\{u_{[\Delta_l:k]},\mathring{u}_{[\Delta_l:k]}\}}(d(u_{[\Delta_l:k]}',\mathring{u}_{[\Delta_l:k]}'))\times \\ & & \mathbf{\check{Q}}_{\theta}^l\Big((u_{k}',\mathring{u}_{k}'),d(u_{[k+\Delta_l:k+1]}',\mathring{u}_{[k+\Delta_l:k+1]}')\Big)
\end{eqnarray*}
where
$$
\mathbf{\check{Q}}_{\theta}^l\Big((u_{k}',\mathring{u}_{k}'),d(u_{[k+\Delta_l:k+1]}',\mathring{u}_{[k+\Delta_l:k+1]}')\Big)= 
$$
\begin{equation}\label{eq:chql_def}
\prod_{m=1}^{\Delta_{l}^{-1}}\check{Q}_{\theta}^l\Big((u_{k+(m-1)\Delta_{l}^{-1}}',\mathring{u}_{k+(m-1)\Delta_{l}^{-1}}'),d(u_{k+m\Delta_{l}^{-1}}',\mathring{u}_{k+m\Delta_{l}^{-1}}')\Big).
\end{equation}
Then we set, with $(x_{[\Delta_l:T]},\mathring{x}_{[\Delta_l:T]})\in\mathsf{X}_l^T\times\mathsf{X}_l^T$, 
$$
\mathbb{K}^l_{T,\theta}\Big((x_{[\Delta_l:T]},\mathring{x}_{[\Delta_l:T]}),d((\mathbf{u}_0^{l,1:N},\mathbf{\mathring{u}}_0^{l,1:N}),\dots,(\mathbf{u}_{T-1}^{l,1:N},\mathbf{\mathring{u}}_{T-1}^{l,1:N}))\Big) = 
$$
$$
\prod_{i=1}^{N-1} \mathbf{\check{Q}}_{\theta}^l\Big((x,x),d(\mathbf{u}_{0}^{l,i},\mathbf{\mathring{u}}_{0}^{l,i})\Big)\delta_{\{x_{[\Delta_l:1]},\mathring{x}_{[\Delta_l:1]}\}}\Big(d(\mathbf{u}_0^{l,N},\mathbf{\mathring{u}}_0^{l,N})\Big)
$$
$$
\Big[\prod_{k=1}^{T-1} 
\Big\{
\prod_{i=1}^{N-1}\sum_{(r,s)\in\{1,\dots,N\}^2}
\omega_{k-1,\theta}^{l}(r,s,\mathbf{u}_{k-1,k-1}^{l,1:N},\mathbf{\mathring{u}}_{k-1,k-1}^{l,1:N})
$$
$$
\mathbf{Q}_{k,\theta}^l\Big((\mathbf{u}_{k-1}^{l,r},\mathbf{\mathring{u}}_{k-1}^{l,s}),d(\mathbf{u}_{k}^{l,i},\mathbf{\mathring{u}}_{k}^{l,i})\Big)
\Big\}\delta_{\{x_{[\Delta_l:k]},\mathring{x}_{[\Delta_l:k]}\}}\Big(d(\mathbf{u}_{k}^{l,N},\mathbf{\mathring{u}}_k^{l,N})\Big)
\Big].\label{eq:p_def_ccpf}
$$

Now the CCPF kernel $K_{\theta}^l:\mathsf{X}_l^{2T}\rightarrow\mathcal{P}(\mathsf{X}_l^{2T})$ is defined as
$$
K_{T,\theta}^l\Big((x_{[\Delta_l:T]},\mathring{x}_{[\Delta_l:T]}),d(x_{[\Delta_l:T]}',\mathring{x}_{[\Delta_l:T]}')\Big) = 
$$
$$
\int_{\mathsf{X}_l^{2NT}} 
\mathbb{K}^l_{T,\theta}\Big((x_{[\Delta_l:T]},\mathring{x}_{[\Delta_l:T]}),d((\mathbf{u}_0^{l,1:N},\mathbf{\mathring{u}}_0^{l,1:N}),\dots,(\mathbf{u}_{T-1}^{l,1:N},\mathbf{\mathring{u}}_{T-1}^{l,1:N}))\Big)
$$
$$
\sum_{(r,s)\in\{1,\dots,N\}^2}
\omega_{T-1,\theta}^{l}(r,s,\mathbf{u}_{T-1,T-1}^{l,1:N},\mathbf{\mathring{u}}_{T-1,T-1}^{l,1:N})\delta_{\{\mathbf{u}_{T-1}^{l,r},\mathbf{\mathring{u}}_{T-1}^{l,s}\}}(d(x_{[\Delta_l:T]}',\mathring{x}_{[\Delta_l:T]}')).
$$
The simulation of the CCPF kernel is described in detail in Algorithm \ref{algo:ccpf}.

\begin{algorithm}[H]
\begin{enumerate}
\item{Initialize: For $i\in\{1,\dots,N-1\}$ sample $\mathbf{u}_0^{l,i},\mathbf{\mathring{u}}_0^{l,i}$ independently using $\mathbf{\check{Q}}_{\theta}^l\Big((x,x),\cdot\Big)$. Set $(\mathbf{u}_0^{l,N},\mathbf{\mathring{u}}_0^{l,N})=(x_{[\Delta_l:1]},\mathring{x}_{[\Delta_l:1]})$, $k=0$.}
\item{Coupled Resampling:  For $i\in\{1,\dots,N-1\}$ sample $\kappa^i\sim\mathcal{U}_{[0,1]}$. If \\
$\kappa^i<\Big(\sum_{s=1}^N \{F_{k,\theta}^l(s,\mathbf{u}_{k,k}^{l,1:N})\wedge
F_{k,\theta}^l(s,\mathring{\mathbf{u}}_{k,k}^{l,1:N})\}\Big)$, then sample $j^i$ from 
$$
\frac{F_{k,\theta}^l(j^i,\mathbf{u}_{k,k}^{l,1:N})\wedge
F_{k,\theta}^l(j^i,\mathring{\mathbf{u}}_{k,k}^{l,1:N})}{\sum_{s=1}^N\{ F_{k,\theta}^l(s,\mathbf{u}_{k,k}^{l,1:N})\wedge
F_{k,\theta}^l(s,\mathring{\mathbf{u}}_{k,k}^{l,1:N})\}}
$$
and set $r_k^i=s_k^i=j^i$. Otherwise, sample $j_1^i$ and $j_2^i$ from 
$$
\Bigg(\frac{F_{k,\theta}^l(j_1^i,\mathbf{u}_{k,k}^{l,1:N})-F_{k,\theta}^l(j_1^i,\mathbf{u}_{k,k}^{l,1:N})\wedge F_{k,\theta}^l(j_1^i,\mathring{\mathbf{u}}_{k,k}^{l,1:N})}{1-\sum_{s=1}^N\{F_{k,\theta}^l(s,\mathbf{u}_{k,k}^{l,1:N})\wedge F_{k,\theta}^l(s,\mathring{\mathbf{u}}_{k,k}^{l,1:N})\}}\Bigg) 
$$
$$
\Bigg(\frac{F_{k,\theta}^l(j_2^i,\mathring{\mathbf{u}}_{k,k}^{l,1:N})-F_{k,\theta}^l(j_2^i,\mathbf{u}_{k,k}^{l,1:N})\wedge F_{k,\theta}^l(j_2^i,\mathring{\mathbf{u}}_{k,k}^{l,1:N})}{1-\sum_{s=1}^N\{F_{k,\theta}^l(s,\mathbf{u}_{k,k}^{l,1:N})\wedge F_{k,\theta}^l(s,\mathring{\mathbf{u}}_{k,k}^{l,1:N})\}}\Bigg)
$$
and set $r_k^i=j_1^i$ and $s_k^i=j_2^i$.
}
\item{Coupled Sampling:  Set $k=k+1$. For $i\in\{1,\dots,N-1\}$ sample $\mathbf{u}_k^{l,i},\mathbf{\mathring{u}}_k^{l,i}| \mathbf{u}_{k-1}^{l,r_{k-1}^i},\mathbf{\mathring{u}}_{k-1}^{l,s_{k-1}^i}$ conditionally independently using $\mathbf{Q}_{k,\theta}^l\Big((\mathbf{u}_{k-1}^{l,r_{k-1}^i},\mathbf{\mathring{u}}_{k-1}^{l,s_{k-1}^i}),\cdot\Big)$. 
Set $(\mathbf{u}_k^{l,N},\mathbf{\mathring{u}}_k^{l,N})=(x_{[\Delta_l:k+1]},\mathring{x}_{[\Delta_l:k+1]})$. If $k=T-1$ go to 4., otherwise go to 2..
}
\item{Select Trajectories: Sample $(r^l,s^l)$ according to $\omega_{T-1,\theta}^{l}(r^l,s^l,\mathbf{u}_{T-1,T-1}^{l,1:N},\mathbf{\mathring{u}}_{T-1,T-1}^{l,1:N})$ (as described, for one
$i$ in 2.).
Return $(\mathbf{u}_{T-1}^{l,r^l},\mathbf{\mathring{u}}_{T-1}^{l,s^l})$.
}
\end{enumerate}
\caption{Simulating the CCPF kernel.}
\label{algo:ccpf}
\end{algorithm}

\subsubsection{Initial distribution}   \label{sec:ccpf_in_distr}

In this subsection, we define an initial distribution $\mu_{T,\theta}^l\in\mathcal{P}(\mathsf{X}_l^{2T})$ that we use to initialize the CCPF $(X_{[\Delta_l:T]}^{(l,0)},\mathring{X}_{[\Delta_l:T]}^{(l,0)}),\dots$, $(X_{[\Delta_l:T]}^{(l,k)},\mathring{X}_{[\Delta_l:T]}^{(l,k)})\in \mathsf{X}_l^{2T}$, $k\in\mathbb{Z}^+$. 
The initialization of the CCPF consists of generating trajectories $X_{[\Delta_l:T]}^{(l,0)}$ and $\overline{X}_{[\Delta_l:T]}^{(l,0)}$ independently using transition kernel $Q^{l}_{\theta}$. Then, we apply the CPF kernel $\overline{K}^{l}_{T,\theta}:\mathsf{X}^T_l \rightarrow \mathcal{P}(\mathsf{X}^T_l)$, as in Section \ref{sec:cpf}, conditional on trajectory $\overline{X}_{[\Delta_l:T]}^{(l,0)}$.
Thus, from the above discussion, it follows that the initial distribution $\mu_{T,\theta}^l\in\mathcal{P}(\mathsf{X}_l^{2T})$ is formalized as
\begin{equation} \label{eq:id_0}
\begin{split}
&\mu_{T,\theta}^l(d(x_{[\Delta_l:T]},\mathring{x}_{[\Delta_l:T]})) =   \\
&\Big(\prod_{k=1}^{T}Q_{\theta}^l(x_{(k-1)},dx_{k})\Big)
\Big(\prod_{k=1}^{T}Q_{\theta}^l(\overline{x}_{(k-1)},d\overline{x}_{k})\Big)\Big(\overline{K}^{l}_{T,\theta}\big(\overline{x}_{[\Delta_l:T]},d\mathring{x}_{[\Delta_l:T]}\Big).  
\end{split}
\end{equation}

\subsubsection{Rhee-Glynn estimator}   \label{sec:rhee-glynn_estimator}

The principle of the CCPF is to use a randomization technique as in \cite{glynn} (see also \cite{rhee,vihola}) to obtain an unbiased estimate of $\pi_{\theta}^l(\varphi_{\theta}^l)$ by simulating a Markov chain $(X_{[\Delta_l:T]}^{(l,0)},\mathring{X}_{[\Delta_l:T]}^{(l,0)}),\dots$
of initial distribution $\mu^{l}_{T,\theta}\in\mathcal{P}(\mathsf{X}_l^{2T})$ and transition kernel $K_{\theta}^l$. Defining the meeting time as $\tau^l = \inf\{k\geq 1:X_{[\Delta_l:T]}^{(l,k)}=\mathring{X}_{[\Delta_l:T]}^{(l,k)}\}$ and setting a $k^\star\in\{2,3,\dots\}$ (the choice of this parameter is discussed in \cite{jacob2}), one generates the Markov chain as described up to time $M=\max(\tau^l,k^\star)$, and considers the estimator
\begin{eqnarray}
\hat{\pi}_\theta^l(\varphi_{\theta}^l) := \varphi_{\theta}^l\Big(X_{[\Delta_l:T]}^{l,k^\star}\Big) + \sum_{k=k^\star+1}^{\tau^{l}-1}\Big\{\varphi_{\theta}^l\Big(X_{[\Delta_l:T]}^{l,k}\Big)-\varphi_{\theta}^l\Big(\mathring{X}_{[\Delta_l:T]}^{l,k}\Big)\Big\},   \label{eq:rhee_glynn}
\end{eqnarray}
with the second term equal to zero if $\tau^l-1\leq k^\star+1$. In \cite{jacob2}, it is demonstrated that under some assumptions, $\hat{\pi}_\theta^l(\varphi_{\theta}^l)$ is an unbiased estimator $\pi_\theta^l(\varphi_{\theta}^l)$.

To improve the variance of \eqref{eq:rhee_glynn}, as described in \cite{jacob2}, we consider the estimator 
\begin{equation}   \label{eq:rhee_glynn_tune}
\begin{split}
\hat{\pi}_\theta^l(\varphi_{\theta}^l) &:= \frac{1}{m^\star-k^\star+1}\sum_{k=k^\star}^{m^\star} \varphi_{\theta}^{l}\Big(X^{l,k}_{[\Delta_l:T]}\Big)+ \\ &\sum_{k=k^\star+1}^{\tau^{l}-1}\frac{\min(m^\star-k^\star+1,k-k^\star)}{m^\star-k^\star+1}\Big(\varphi_{\theta}^l\Big(X_{[\Delta_l:T]}^{l,k}\Big)-\varphi_{\theta}^l\Big(\mathring{X}_{[\Delta_l:T]}^{l,k}\Big)\Big),
\end{split}
\end{equation}
where $k^\star<m^\star$. The first term on the left-hand side consists of an average between $k^\star$ and $m^\star$ of the Markov chain.

To further reduce the variance of the proposed estimators, we evaluate $\varphi_{\theta}^{l}\Big(X^{l}_{[\Delta_{l}:T]}\Big)$ over $N$ trajectories $u^{l,1:N}_{[\Delta_{l}:T]}$ simulating the CCPF kernel, such that it becomes 
$$
\varphi_{\theta}^{l}\Big(X^{l}_{[\Delta_{l}:T]}\Big)=\sum_{s=1}^{N}F^{l}_{T-1,\theta}(s,u^{l,1:N}_{T-1,T-1})\varphi_{\theta}^{l}(u^{l,s}_{T-1}).
$$
This estimator has the same expectation as $\varphi_{\theta}^{l}(X^{l,k^\star}_{[\Delta_l:T]})$. The same procedure can be applied to compute $\varphi_{\theta}^{l}\Big(\mathring{X}^{l}_{[\Delta_{l}:T]}\Big)$.

\subsection{Coupling of CCPF (C-CCPF)}\label{sec:cccpf}

Throughout this section $l\geq 1$ and $\theta\in\Theta$ are both fixed. 

\subsubsection{Probability Kernel Coupling}\label{sec:cccpf_pkc}

We now introduce a Markov kernel $\mathbf{\check{Q}}_{\theta}^{l,l-1}:\mathbb{R}^{4d_x}\rightarrow\mathcal{P}(\mathsf{X}_l^2\times\mathsf{X}_{l-1}^2)$ whose simulation is described in Algorithm \ref{algo:qll_sim}. Given the description, it can be easily verified that for any 
$\Big((u_0^l,\mathring{u}_0^l),(u_0^{l-1},\mathring{u}_0^{l-1})\Big)\in\mathbb{R}^{2d_x}\times\mathbb{R}^{2d_x}$ and any $(A,\mathring{A})\in B(\mathsf{X}_l)\vee B(\mathsf{X}_{l-1})$,
\begin{eqnarray*}
\mathbf{\check{Q}}_{\theta}^{l,l-1}(A\times\mathsf{X}_{l-1})\Big((u_0^l,\mathring{u}_0^l),(u_0^{l-1},\mathring{u}_0^{l-1})\Big) &  = &
\mathbf{\check{Q}}_{\theta}^{l}(A)\Big((u_0^l,\mathring{u}_0^l)\Big) \\
\mathbf{\check{Q}}_{\theta}^{l,l-1}(\mathsf{X}_{l}\times\mathring{A})\Big((u_0^l,\mathring{u}_0^l),(u_0^{l-1},\mathring{u}_0^{l-1})\Big) & = &
\mathbf{\check{Q}}_{\theta}^{l-1}(\mathring{A})\Big((u_0^{l-1},\mathring{u}_0^{l-1})\Big),
\end{eqnarray*}
where $\mathbf{\check{Q}}_{\theta}^{l}$ and $\mathbf{\check{Q}}_{\theta}^{l-1}$ are as detailed \eqref{eq:chql_def}.

\begin{algorithm}[H]
\begin{enumerate}
\item{Input $\Big((u_0^l,\mathring{u}_0^l),(u_0^{l-1},\mathring{u}_0^{l-1})\Big)\in\mathbb{R}^{2d_x}\times\mathbb{R}^{2d_x}$.}
\item{Generate $W_k\stackrel{i.i.d.}{\sim}\mathcal{N}_{d_x}(0,\Delta_l I_{d_x})$, $k\in\{1,2\dots,\Delta_l^{-1}\}$.}
\item{For $k\in\{1,2\dots,\Delta_l^{-1}\}$:
\begin{eqnarray*}
U_{k\Delta_l}^l & = & U_{(k-1)\Delta_l}^l + b_{\theta}(U_{(k-1)\Delta_l}^l)\Delta_l + \sigma(U_{(k-1)\Delta_l}^l)W_k \\
\mathring{U}_{k\Delta_l}^l & = & \mathring{U}_{(k-1)\Delta_l}^l + b_{\theta}(\mathring{U}_{(k-1)\Delta_l}^l)\Delta_l + \sigma(\mathring{U}_{(k-1)\Delta_l}^l)W_k
\end{eqnarray*}
}
\item{For $k\in\{1,2\dots,\Delta_{l-1}^{-1}\}$:
\begin{eqnarray*}
U_{k\Delta_{l-1}}^{l-1} & = & U_{(k-1)\Delta_{l-1}}^{l-1} + b_{\theta}(U_{(k-1)\Delta_{l-1}}^{l-1})\Delta_{l-1} + \sigma(U_{(k-1)\Delta_{l-1}}^{l-1})[W_{2k-1}+W_{2k}] \\
\mathring{U}_{k\Delta_{l-1}}^{l-1} & = & \mathring{U}_{(k-1)\Delta_{l-1}}^{l-1} + b_{\theta}(\mathring{U}_{(k-1)\Delta_{l-1}}^{l-1})\Delta_{l-1} + \sigma(\mathring{U}_{(k-1)\Delta_{l-1}}^{l-1})[W_{2k-1}+W_{2k}]
\end{eqnarray*}
}
\item{Output $\Big((u_{\Delta_l}^l,\mathring{u}_{\Delta_l}^l),\dots,(u_{1}^l,\mathring{u}_{1}^l)\Big)\in\mathsf{X}_l^2$ and $\Big((u_{\Delta_{l-1}}^{l-1},\mathring{u}_{\Delta_{l-1}}^{l-1}),\dots,(u_{1}^{l-1},\mathring{u}_{1}^{l-1})\Big)\in\mathsf{X}_{l-1}^2$.}
\end{enumerate}
\caption{Simulating $\mathbf{\check{Q}}_{\theta}^{l,l-1}$.}
\label{algo:qll_sim}
\end{algorithm}

\subsubsection{C-CCPF Kernel}\label{sec:cccpf_kernel}

Now define the probability kernel, for $k\in\{1,\dots,T-1\}$, $\mathbf{\check{Q}}_{k,\theta}^{l,l-1}:\mathsf{X}_l^{2k}\times\mathsf{X}_{l-1}^{2k}\rightarrow\mathcal{P}(\mathsf{X}_l^{2(k+1)}\times\mathsf{X}_{l-1}^{2(k+1)})$, $
(v_{[\Delta_l:k]}^l,v_{[\Delta_{l-1}:k]}^{l-1}):=\Big((u_{[\Delta_l:k]}^l,\mathring{u}^l_{[\Delta_l:k]}),(u_{[\Delta_{l-1}:k]}^{l-1},\mathring{u}_{[\Delta_{l-1}:k]}^{l-1})\Big)\in\mathsf{X}_l^{2k}\times\mathsf{X}_{l-1}^{2k}$
\begin{eqnarray*}
\begin{split}
& \mathbf{\check{Q}}_{k,\theta}^{l,l-1}\Big((v_{[\Delta_{l}:k]}^l,v_{[\Delta_{l-1}:k]}^{l-1}),d(v_{[\Delta_{l}:k+1]}^{l,'},v_{[\Delta_{l-1}:k+1]}^{l-1,'})\Big) := \\
&\delta_{\{v_{[\Delta_{l}:k]}^l,v_{[\Delta_{l-1}:k]}^{l-1}\}}(d(v_{[\Delta_{l}:k]}^{l,'},v_{[\Delta_{l-1}:k]}^{l-1,'}))\times  
\mathbf{\check{Q}}_{\theta}^{l,l-1}\Big((v_{k}^{l,'},v_{k}^{l-1,'}),d(v_{[k+\Delta_l:k+1]}^{l,'},v_{[k+\Delta_{l-1}:k+1]}^{l-1,'})\Big).
\end{split}
\end{eqnarray*}

We now introduce some additional conventions. Set $v_{[k+\Delta_s,k+1]}^{s,1:N}=(u_{[k+\Delta_s:k+1]}^{s,1:N},\mathring{u}_{[k+\Delta_s:k+1]}^{s,1:N})\in\mathsf{X}_s^{2N}$, $s\in\{l-1,l\}$, $k\in\{0,1\dots,T-1\}$. Let for $s\in\{l-1,l\}$,
$i\in\{1,\dots,N\}$, $k\in\{0,1,\dots,T-1\}$
\begin{eqnarray*}
\mathbf{v}_k^{s,i} & = & (\mathbf{u}_k^{s,i},\mathbf{\mathring{u}}_k^{s,i}) \in \mathsf{X}_s^{2(k+1)}  \\
\mathbf{u}_k^{s,i} & = & (\mathbf{u}_{k,0}^{s,i},\dots,\mathbf{u}_{k,k}^{s,i}) \\
\mathbf{\mathring{u}}_k^{s,i}&=&(\mathbf{\mathring{u}}_{k,0}^{s,i},\dots,\mathbf{\mathring{u}}_{k,k}^{s,i})
\end{eqnarray*}
where $(\mathbf{u}_{k,j}^{s,i},\mathbf{\mathring{u}}_{k,j}^{s,i})\in\mathsf{X}_s^2$, $j\in\{0,1,\dots,k\}$. For $i\in\{1,\dots,N\}$, $k\in\{0,\dots,T-1\}$, we compute quantities $F_{k,\theta}^{l}(i,\mathbf{u}_{k,k}^{l,1:N})$, $F_{k,\theta}^{l}(i,\mathring{\mathbf{u}}_{k,k}^{l,1:N})$, $F_{k,\theta}^{l-1}(i,\mathbf{u}_{k,k}^{l-1,1:N})$ and $F_{k,\theta}^{l-1}(i,\mathring{\mathbf{u}}_{k,k}^{l-1,1:N})$. Quantity $\check{\omega}_{k,\theta}^{l,l-1}\Big((i^l,i^{l-1},j^l,j^{l-1}),\mathbf{v}_{k,k}^{l,1:N},\mathbf{v}_{k,k}^{l-1,1:N}\Big)$ is associated with the Maximal Coupling of Maximal Couplings as described in Algorithm \ref{alg:thorrison}.

\begin{algorithm}[H]
\begin{enumerate}
\item Input: Four Probability Functions $F_{k,\theta}^{l}(i,\mathbf{u}_{k,k}^{l,1:N})$, $F_{k,\theta}^{l}(i,\mathring{\mathbf{u}}_{k,k}^{l,1:N})$, $F_{k,\theta}^{l-1}(i,\mathbf{u}_{k,k}^{l-1,1:N})$, $F_{k,\theta}^{l-1}(i,\mathring{\mathbf{u}}_{k,k}^{l-1,1:N})$
\item{Sample two indices from maximal coupling probability $\omega^{l}_{k,\theta}(r^{l},s^{l},u^{l,1:N}_{[k:k+1]},\mathring{u}^{l,1:N}_{[k:k+1]})$ and evaluate $\omega^{l-1}_{k,\theta}(r^l,s^l,u^{l-1,1:N}_{[k:k+1]},\mathring{u}^{l-1,1:N}_{[k:k+1]})$}
\item Sample $U\sim\mathcal{U}_{[0,\omega^{l}_{k,\theta}(r^{l},s^{l},u^{l,1:N}_{[k:k+1]},\mathring{u}^{l,1:N}_{[k:k+1]})]}$ and if $U<\omega^{l-1}_{k,\theta}(r^l,s^l,u^{l-1,1:N}_{[k:k+1]},\mathring{u}^{l-1,1:N}_{[k:k+1]})$ then return $i^{l}=r^l$, $i^{l-1}=r^{l}$, $j^{l}=s^{l}$, $j^{l-1}=s^{l}$. Otherwise move step 4.
\item Sample two indices from maximal coupling probability $\omega^{l-1}_{k,\theta}(r^{l-1},s^{l-1},u^{l-1,1:N}_{[k:k+1]},\mathring{u}^{l-1,1:N}_{[k:k+1]})$ and evaluate $\omega^{l}_{k,\theta}(r^{l-1},s^{l-1},u^{l,1:N}_{[k:k+1]},\mathring{u}^{l,1:N}_{[k:k+1]})$
\item Sample $U\sim\mathcal{U}_{[0,\omega^{l-1}_{k,\theta}(r^{l-1},s^{l-1},u^{l-1,1:N}_{[k:k+1]},\mathring{u}^{l-1,1:N}_{[k:k+1]})]}$ and if $U>\omega^{l}_{k,\theta}(r^{l-1},s^{l-1},u^{l,1:N}_{[k:k+1]},\mathring{u}^{l,1:N}_{[k:k+1]})$ then return $i^{l}=r^{l}$,$j^{l}=s^{l}$,$i^{l-1}=r^{l-1}$, $j^{l-1}=s^{l-1}$. Otherwise return Step 4.
\end{enumerate}
\caption{Simulating a Maximal Coupling of Maximal Couplings $\check{\omega}_{k,\theta}^{l,l-1}\Big((i^l,i^{l-1},j^l,j^{l-1}),\mathbf{v}_{k,k}^{l,1:N},\mathbf{v}_{k,k}^{l-1,1:N}\Big)$}
\label{alg:thorrison}
\end{algorithm}

As for the CCPF, we introduce an underlying kernel $\check{\mathbb{K}}^{l,l-1}_{T,\theta}:\mathsf{X}_l^{2T}\times\mathsf{X}_{l-1}^{2T}\rightarrow\mathcal{P}(\mathsf{X}_l^{2T}\times \mathsf{X}_{l-1}^{2T})$ which is critical in defining a C-CCPF Markov kernel. Set for $v^s_{[\Delta_s:T]}=(x_{[\Delta_s:T]}^s,\mathring{x}_{[\Delta_s:T]}^s)\in\mathsf{X}_s^{2T}$, $s\in\{l-1,l\}$, $v=(x_*,x_*)$
$$
\check{\mathbb{K}}^{l,l-1}_{T,\theta}\Big((v^l_{[\Delta_{l}:T]},v^{l-1}_{[\Delta_{l-1}:T]}),d((\mathbf{v}_0^{l,1:N},\mathbf{v}_0^{l-1,1:N}),\dots,(\mathbf{v}_{T-1}^{l,1:N},\mathbf{v}_{T-1}^{l-1,1:N}))\Big) = 
$$
$$
\Big\{\prod_{i=1}^{N-1} \mathbf{\check{Q}}_{\theta}^{l,l-1}\Big((v,v),d(\mathbf{v}_0^{l,i},\mathbf{v}_0^{l-1,i})\Big)\Big\}\delta_{\{(v_{[\Delta_{l}:1]}^l,v_{[\Delta_{l-1}:1]}^{l-1})\}}\Big(d(\mathbf{v}_0^{l,N},\mathbf{v}_0^{l-1,N})\Big)
$$
$$
\Big[\prod_{k=1}^{T-1}\Big\{\prod_{i=1}^{N-1}\sum_{(r^l,s^l,r^{l-1},s^{l-1})\in\{1,\dots,N\}^4}
\check{\omega}_{k-1,\theta}^{l,l-1}\Big((r^l,s^l,r^{l-1},s^{l-1}),\mathbf{v}_{k-1,k-1}^{l,1:N},\mathbf{v}_{k-1,k-1}^{l-1,1:N}\Big)\times
$$
$$
\mathbf{\check{Q}}_{k,\theta}^{l,l-1}\Big([(\mathbf{u}_{k-1}^{l,r^l},\mathbf{\mathring{u}}_{k-1}^{l,s^l}),(\mathbf{u}_{k-1}^{l-1,r^{l-1}},\mathbf{\mathring{u}}_{k-1}^{l-1,s^{l-1}})],d(\mathbf{v}_{k}^{l,i},\mathbf{v}_{k}^{l,i})\Big)
\Big\}
\delta_{\{(v_{[\Delta_{l}:k]}^l,v_{[\Delta_{l-1}:k]}^{l-1})\}}\Big(d(\mathbf{v}_{k}^{l,N},\mathbf{v}_k^{l-1,N})\Big)
\Big].
$$
Now the C-CCPF kernel $\check{K}_{\theta}^{l,l-1}:\mathsf{X}_l^{2T}\times\mathsf{X}_{l-1}^{2T}\rightarrow\mathcal{P}(\mathsf{X}_l^{2T}\times\mathsf{X}_{l-1}^{2T})$ is defined as
$$
\check{K}_{T,\theta}^{l,l-1}\Big((v^l_{[\Delta_{l}:T]},v^{l-1}_{[\Delta_{l-1}:T]}),d(v^{l,'}_{[\Delta_{l}:T]},v^{l-1,'}_{[\Delta_{l-1}:T]})\Big) := 
$$
$$
\int_{\mathsf{X}_l^{2NT}\times \mathsf{X}_{l-1}^{2NT}} 
\check{\mathbb{K}}^{l,l-1}_{T,\theta}\Big((v^l_{[\Delta_{l}:T]},v^{l-1}_{[\Delta_{l-1}:T]}),d((\mathbf{v}_0^{l,1:N},\mathbf{v}_0^{l-1,1:N}),\dots,(\mathbf{v}_{T-1}^{l,1:N},\mathbf{v}_{T-1}^{l-1,1:N}))\Big)
$$
$$
\sum_{(r^l,s^l,r^{l-1},s^{l-1})\in\{1,\dots,N\}^4}
\check{\omega}_{n,\theta}^{l,l-1}\Big((r^l,s^l,r^{l-1},s^{l-1}),\mathbf{v}_{n,n}^{l,1:N},\mathbf{v}_{n,n}^{l-1,1:N}\Big)
$$
$$
\delta_{\{
(\mathbf{u}_{n}^{l,r^l},\mathbf{\mathring{u}}_{n}^{l,s^l}),(\mathbf{u}_{n}^{l-1,r^{l-1}},\mathbf{\mathring{u}}_{n}^{l-1,s^{l-1}})
\}}(d(v^{l,'}_{[\Delta_{l}:T]},v^{l-1,'}_{[\Delta_{l-1}:T]})).
$$
The simulation of the C-CCPF kernel is described in Algorithm \ref{algo:cccpf}.

\begin{algorithm}[H]
\begin{enumerate}
\item{Initialize: For $i\in\{1,\dots,N-1\}$ sample $\mathbf{v}_0^{l,i},\mathbf{v}_0^{l-1,i}$ independently using 
$\mathbf{\check{Q}}_{\theta}^{l,l-1}\Big((v,v),\cdot\Big)$. Set $(\mathbf{v}_0^{l,N},\mathbf{v}_0^{l-1,N})=(v_{[\Delta_{l}:1]}^l,v_{[\Delta_{l-1}:1]}^{l-1})$, $k=0$.}
\item{Coupled Resampling $r_k^{l,i}, s_k^{l,i}, r_k^{l-1,i}, s_k^{l-1,i}$ applying Maximal Coupling of Maximal Couplings as Algorithm \ref{alg:thorrison}
}
\item{Coupled Sampling: Set $k=k+1$. For $i\in\{1,\dots,N-1\}$ sample $\mathbf{v}_{k}^{l,i},\mathbf{v}_{k}^{l-1,i}| (\mathbf{u}_{k-1}^{l,r_{k-1}^{l,i}},\mathbf{\mathring{u}}_{k-1}^{l,s_{k-1}^{l,i}}),(\mathbf{u}_{k-1}^{l-1,r_{k-1}^{l-1,i}},\mathbf{\mathring{u}}_{k-1}^{l-1,s_{k-1}^{l-1,i}})$ conditionally independently using $$
\mathbf{\check{Q}}_{k,\theta}^{l,l-1}\Big([(\mathbf{u}_{k-1}^{l,r_{k-1}^{l,i}},\mathbf{\mathring{u}}_{k-1}^{l,s_{k-1}^{l,i}}),(\mathbf{u}_{k-1}^{l-1,r_{k-1}^{l-1.i}},\mathbf{\mathring{u}}_{k-1}^{l-1,s_{k-1}^{l-1.i}})],\cdot\Big).$$
Set $(\mathbf{v}_{k}^{l,N},\mathbf{v}_k^{l-1,N})=((x_{[\Delta_{l}:k+1]}^l,\mathring{x}_{[\Delta_{l}:k+1]}^l),(x_{[\Delta_{l-1}:k+1]}^{l-1},\mathring{x}_{[\Delta_{l-1}:k+1]}^{l-1}))$. If $k=T-1$ go to 4., otherwise go to 2..}
\item{Select Trajectories:  Sample $(r^l,s^l,s^{l-1},s^{l-1})$ according to $\check{\omega}_{n,\theta}^{l,l-1}\Big((r^l,s^l,r^{l-1},s^{l-1}),\mathbf{v}_{n,n}^{l,1:N},\mathbf{v}_{n,n}^{l-1,1:N}\Big)$ (as described, for one
$i$ in 2.).
Return $(\mathbf{u}_{T-1}^{l,r^l},\mathbf{\mathring{u}}_{T-1}^{l,s^l}),(\mathbf{u}_{T-1}^{l-1,r^{l-1}},\mathbf{\mathring{u}}_{T-1}^{l-1,s^{l-1}})$.}
\end{enumerate}
\caption{Simulating the C-CCPF kernel. }
\label{algo:cccpf}
\end{algorithm}

\subsubsection{Initial Distribution} \label{sec:cccpf_in_distr}
For any $l\geq 1$, we simulate a Markov chain $(V_{[\Delta_{l}:T]}^{(l,0)},V_{[\Delta_{l-1}:T]}^{(l-1,0)}),\dots$, $(V_{[\Delta_{l}:T]}^{(l,k)},V_{[\Delta_{l-1}:T]}^{(l-1,k)})\in \mathsf{X}_l^{2T}\times \mathsf{X}_{l-1}^{2T}$, $k\in\mathbb{Z}^+$ of initial distribution $\mu_{T,\theta}^{l,l-1}\in\mathcal{P}(\mathsf{X}_l^{2T}\times \mathsf{X}_{l-1}^{2T})$, with
$\Big((x_0^l,\mathring{x}_0^{l}),(x_0^{l-1},\mathring{x}_0^{l-1})\Big)=\Big((x_*,x_*),(x_*,x_*)\Big)$.
The initial distribution consists of generating two pairs of coupled trajectories: $\big(X^{(l,0)}_{[\Delta_l:T]}, X^{(l-1,0)}_{[\Delta_{l-1}:T]}\big)$ and $\big(\mathring{X}^{(l,0)}_{[\Delta_l:T]}, \mathring{X}^{(l-1,0)}_{[\Delta_{l-1}:T]}\big)$. We build the first coupled trajectories $\big(X^{(l,0)}_{[\Delta_l:T]}, X^{(l-1,0)}_{[\Delta_{l-1}:T]}\big)$ by the CCPF kernel $\overline{K}^{l,l-1}_{\theta}:\mathsf{X}^{T}_{l}\times \mathsf{X}^{T}_{l-1}\rightarrow \mathcal{P}(\mathsf{X}^{T}_{l}\times \mathsf{X}^{T}_{l-1})$, defined as 
$$
\overline{K}_{T,\theta}^{l,l-1}\Big((x_{[\Delta_l:T]},x_{[\Delta_{l-1}:T]}),d(x_{[\Delta_l:T]}',x_{[\Delta_{l-1}:T]}')\Big) = 
$$
$$
\int_{\mathsf{X}_l^{NT}\times\mathsf{X}_{l-1}^{NT}} 
\overline{\mathbb{K}}^{l,l-1}_{T,\theta}\Big((x_{[\Delta_l:T]},x_{[\Delta_{l-1}:T]}),d((\mathbf{u}_0^{l,1:N},\mathbf{u}_0^{l-1,1:N}),\dots,(\mathbf{u}_{T-1}^{l,1:N},\mathbf{u}_{T-1}^{l-1,1:N}))\Big)
$$
$$
\sum_{(r,s)\in\{1,\dots,N\}^2}\omega_{T-1,\theta}^{l,l-1}(r,s,\mathbf{u}_{T-1,T-1}^{l,1:N},\mathbf{u}_{T-1,T-1}^{l-1,1:N})\delta_{\{\mathbf{u}_{T-1}^{l,r},\mathbf{u}_{T-1}^{l-1,s}\}}(d(x_{[\Delta_l:T]}',x_{[\Delta_{l-1}:T]}'))
$$
where, with $(x_{[\Delta_l:T]},x_{[\Delta_{l-1}:T]})\in\mathsf{X}^{T}_{l}\times\mathsf{X}^{T}_{l-1}$,
$$
\overline{\mathbb{K}}^{l,l-1}_{T,\theta}\Big((x_{[\Delta_l:T]},x_{[\Delta_{l-1}:T]}),d((\mathbf{u}_0^{l,1:N},\mathbf{u}_0^{l-1,1:N}),\dots,(\mathbf{u}_{T-1}^{l,1:N},\mathbf{u}_{T-1}^{l-1,1:N}))\Big) = 
$$
$$
\prod_{i=1}^{N-1} \mathbf{Q}_{\theta}^{l,l-1}\Big((x,x),d(\mathbf{u}_{0}^{l,i},\mathbf{u}_{0}^{l-1,i})\Big)\delta_{\{x_{[\Delta_l:1]},x_{[\Delta_{l-1}:1]}\}}\Big(d(\mathbf{u}_0^{l,N},\mathbf{u}_0^{l-1,N})\Big)
$$
$$
\Big[\prod_{k=1}^{T-1} 
\Big\{
\prod_{i=1}^{N-1}\sum_{(r,s)\in\{1,\dots,N\}^2}
\omega_{k-1,\theta}^{l,l-1}(r,s,\mathbf{u}_{k-1,k-1}^{l,1:N},\mathbf{u}_{k-1,k-1}^{l-1,1:N})\times
$$
$$
\overline{\mathbf{Q}}_{k,\theta}^{l,l-1}\Big((\mathbf{u}_{k-1}^{l,r},\mathbf{u}_{k-1}^{l-1,s}),d(\mathbf{u}_{k}^{l,i},\mathbf{u}_{k}^{l-1,i})\Big)
\Big\}\delta_{\{x_{[\Delta_l:k]},x_{[\Delta_{l-1}:k]}\}}\Big(d(\mathbf{u}_{k}^{l,N},\mathbf{u}_k^{l-1,N})\Big)
\Big]\label{eq:p_def_ccpf_l_l1}
$$
with probability kernel, for $k\in\{1,\dots,T-1\}$, $\overline{\mathbf{Q}}^{l,l-1}_{k,\theta}:\mathsf{X}^{k}_{l}\times\mathsf{X}^{k}_{l-1}\rightarrow\mathcal{P}(\mathsf{X}^{(k+1)}_{l}\times\mathsf{X}^{(k+1)}_{l-1})$, $(u^{l}_{[\Delta_l:k]},u^{l-1}_{[\Delta_{l-1}:k]})\in\mathsf{X}^{k}_{l}\times\mathsf{X}^{k}_{l-1}$
\begin{eqnarray*}
\begin{split}
&\overline{\mathbf{Q}}_{k,\theta}^{l,l-1}\Big((u_{[\Delta_{l}:k]}^l,u_{[\Delta_{l-1}:k]}^{l-1}),d(u_{[\Delta_{l}:k+1]}^{l,'},u_{[\Delta_{l-1}:k+1]}^{l-1,'})\Big) :=  \\
&\delta_{\{
u_{[\Delta_{l}:k]}^l,u_{[\Delta_{l-1}:k]}^{l-1}\}}(d(u_{[\Delta_{l}:k]}^{l,'},u_{[\Delta_{l-1}:k]}^{l-1,'}))\times  
\mathbf{Q}_{\theta}^{l,l-1}\Big((u_{k}^{l,'},u_{k}^{l-1,'}),d(u_{[k+\Delta_l:k+1]}^{l,'},u_{[k+\Delta_{l-1}:k+1]}^{l-1,'})\Big),
\end{split}
\end{eqnarray*}
with Markov kernel $\mathbf{Q}_{\theta}^{l,l-1}:\mathbf{R}^{2d_x}\rightarrow\mathcal{P}(\mathsf{X}_{l}\times\mathsf{X}_{l-1})$, whose simulation is described in Algorithm \ref{algo:qll_level_sim}. Finally, $\omega^{l,l-1}(i,j,u^{l,1:N}_{[k+\Delta_l:k+1]},u^{l-1}_{[k+\Delta_{l-1}:k+1]})$ corresponds exactly to the maximum coupling of the previously described resampling probabilities.
Kernel $\overline{K}^{l,l-1}_{T,\theta}$ is a CCPF kernel with coupled trajectories on level $l$ and level $l-1$. The algorithm is described in Algorithm \ref{algo:ccpf_level}.

The second pair of trajectories $(\mathring{X}^{(l,0)}_{[\Delta_l:T]}, \mathring{X}^{(l-1,0)}_{[\Delta_{l-1}:T]})$ is simply coupled by probability kernel $\mathbf{Q}_{\theta}^{l,l-1}:\mathbf{R}^{2d_x}\rightarrow\mathcal{P}(\mathsf{X}_{l}\times\mathsf{X}_{l-1})$.

Thus, we define the initial distribution $\mu_{T,\theta}^{l,l-1}\in\mathcal{P}(\mathsf{X}_l^{2T}\times \mathsf{X}_{l-1}^{2T})$ as
\begin{eqnarray}
\mu_{T,\theta}^{l,l-1}(d(v^{l}_{[\Delta_l:T]},v^{l-1}_{[\Delta_{l-1}:T]})) &=& \Bigg\{\prod_{k=1}^{T}\mathbf{Q}_{\theta}^{l,l-1}\Big((u_{k-1}^{l},u_{k-1}^{l-1}),d(u^{l}_{k-1+\Delta_l:k},u^{l-1}_{k-1+\Delta_{l-1}:k})\Big)\Bigg\} \nonumber \\
&\times& \Bigg\{\prod_{k=1}^{T}\mathbf{Q}_{\theta}^{l,l-1}\Big((\overline{u}_{k-1}^{l},\overline{u}_{k-1}^{l-1}),d(\overline{u}^{l}_{k-1+\Delta_l:k},\overline{u}^{l-1}_{k-1+\Delta_{l-1}:k})\Big) \nonumber \\
&\times& \overline{K}_{T,\theta}^{l,l-1}\Big((\overline{x}_{[\Delta_l:T]},\overline{x}_{[\Delta_{l-1}:T]}),d(\mathring{x}_{[\Delta_l:T]}',\mathring{x}_{[\Delta_{l-1}:T]}')\Big)\Bigg\}  \label{eq:id_l_l1}
\end{eqnarray}

\begin{algorithm}
\begin{enumerate}
\item{Input $(u_0^l,u_0^{l-1})\in\mathbb{R}^{d_x}\times\mathbb{R}^{d_x}$.}
\item{Generate $W_k\stackrel{i.i.d.}{\sim}\mathcal{N}_{d_x}(0,\Delta_l I_{d_x})$, $k\in\{1,2\dots,\Delta_l^{-1}\}$.}
\item{For $k\in\{1,2\dots,\Delta_l^{-1}\}$:
\begin{eqnarray*}
U_{k\Delta_l}^l & = & U_{(k-1)\Delta_l}^l + b_{\theta}(U_{(k-1)\Delta_l}^l)\Delta_l + \sigma(U_{(k-1)\Delta_l}^l)W_k 
\end{eqnarray*}
}
\item{For $k\in\{1,2\dots,\Delta_{l-1}^{-1}\}$:
\begin{eqnarray*}
U_{k\Delta_{l-1}}^{l-1} & = & U_{(k-1)\Delta_{l-1}}^{l-1} + b_{\theta}(U_{(k-1)\Delta_{l-1}}^{l-1})\Delta_{l-1} + \sigma(U_{(k-1)\Delta_{l-1}}^{l-1})[W_{2k-1}+W_{2k}] 
\end{eqnarray*}
}
\item{Output $(u_{\Delta_l}^l,\dots,u_{1}^l)\in\mathsf{X}_l$ and $(u_{\Delta_{l-1}}^{l-1},\dots,u_{1}^{l-1})\in\mathsf{X}_{l-1}$.}
\end{enumerate}
\caption{Simulating $\mathbf{Q}_{\theta}^{l,l-1}$.}
\label{algo:qll_level_sim}
\end{algorithm}

\begin{algorithm}
\begin{enumerate}
\item{Initialize: For $i\in\{1,\dots,N-1\}$ sample $\mathbf{u}_0^{l,i},\mathbf{\mathring{u}}_0^{l-1,i}$ independently using $\mathbf{Q}_{\theta}^{l,l-1}\Big((x,x),\cdot\Big)$. Set $(\mathbf{u}_0^{l,N},\mathbf{u}_0^{l-1,N})=(x_{[\Delta_l:1]},x_{[\Delta_{l-1}:1]})$, $k=0$.}
\item{Coupled Resampling:  For $i\in\{1,\dots,N-1\}$ sample $\kappa^i\sim\mathcal{U}_{[0,1]}$. If $\kappa^i<\Big(\sum_{s=1}^N \{F_{k,\theta}^l(s,\mathbf{u}_{k,k}^{l,1:N})\wedge
F_{k,\theta}^{l-1}(s,\mathring{\mathbf{u}}_{k,k}^{l,1:N})\}\Big)$, then sample $j^i$ from 
$$
\frac{F_{k,\theta}^l(j^i,\mathbf{u}_{k,k}^{l,1:N})\wedge
F_{k,\theta}^{l-1}(j^i,\mathbf{u}_{k,k}^{l-1,1:N})}{\sum_{s=1}^N\{ F_{k,\theta}^{l}(s,\mathbf{u}_{k,k}^{l,1:N})\wedge
F_{k,\theta}^{l-1}(s,\mathbf{u}_{k,k}^{l-1,1:N})\}}
$$
and set $r_k^i=s_k^i=j^i$. Otherwise, sample $j_1^i$ and $j_2^i$ from 
$$
\Bigg(\frac{F_{k,\theta}^l(j_1^i,\mathbf{u}_{k,k}^{l,1:N})-F_{k,\theta}^{l}(j_1^i,\mathbf{u}_{k,k}^{l,1:N})\wedge F_{k,\theta}^{l-1}(j_1^i,\mathbf{u}_{k,k}^{l-1,1:N})}{1-\sum_{s=1}^N\{F_{k,\theta}^l(s,\mathbf{u}_{k,k}^{l,1:N})\wedge F_{k,\theta}^{l-1}(s,\mathbf{u}_{k,k}^{l-1,1:N})\}}\Bigg)
$$
$$
\Bigg(\frac{F_{k,\theta}^{l-1}(j_2^i,\mathbf{u}_{k,k}^{l-1,1:N})-F_{k,\theta}^l(j_2^i,\mathbf{u}_{k,k}^{l,1:N})\wedge F_{k,\theta}^{l-1}(j_2^i,\mathbf{u}_{k,k}^{l-1,1:N})}{1-\sum_{s=1}^N\{F_{k,\theta}^l(s,\mathbf{u}_{k,k}^{l,1:N})\wedge F_{k,\theta}^{l-1}(s,\mathbf{u}_{k,k}^{l-1,1:N})\}}\Bigg)
$$
and set $r_k^i=j_1^i$ and $s_k^i=j_2^i$.
}
\item{Coupled Sampling:  Set $k=k+1$. For $i\in\{1,\dots,N-1\}$ sample $\mathbf{u}_k^{l,i},\mathbf{u}_k^{l-1,i}| \mathbf{u}_{k-1}^{l,r_{k-1}^i},\mathbf{u}_{k-1}^{l-1,s_{k-1}^i}$ conditionally independently using $\mathbf{Q}_{k,\theta}^{l,l-1}\Big((\mathbf{u}_{k-1}^{l,r_{k-1}^i},\mathbf{u}_{k-1}^{l-1,s_{k-1}^i}),\cdot\Big)$. 
Set $(\mathbf{u}_k^{l,N},\mathbf{u}_k^{l-1,N})=(x_{[\Delta_l:k+1]},x_{[\Delta_{l-1}:k+1]})$. If $k=T-1$ go to 4., otherwise go to 2..
}
\item{Select Trajectories: Sample $(r^l,s^{l-1})$ according to $\omega_{T-1,\theta}^{l,l-1}(r^l,s^{l-1},\mathbf{u}_{T-1,T-1}^{l,1:N},\mathbf{u}_{T-1,T-1}^{l-1,1:N})$ (as described, for one
$i$ in 2.).
Return $(\mathbf{u}_{T-1}^{l,r^l},\mathbf{u}_{T-1}^{l-1,s^{l-1}})$.
}
\end{enumerate}
\caption{Simulating the CCPF kernel at level $l$,$l-1$, $l\in\mathbb{N}$.}
\label{algo:ccpf_level}
\end{algorithm}

\subsection{Estimate}  \label{sec:cccpf_estimate}

We now describe, on the basis of the approaches presented in Section \ref{sec:ccpf}-\ref{sec:cccpf}, how to construct the random variables $\Psi_{T,\theta}^0,\Psi_{T,\theta}^1,\dots$ from Section \ref{sec:main_idea} to compute an almost-sure unbiased estimate of the gradient of the log-likelihood \eqref{eq:gll}.

To construct $\Psi_{T,\theta}^0$, we simulate a Markov chain $(X_{[\Delta_0:T]}^{(0,0)},\mathring{X}_{[\Delta_0:T]}^{(0,0)}),\dots$, $(X_{[\Delta_0:T]}^{(0,k)},\mathring{X}_{[\Delta_0:T]}^{(0,k)})\in \mathsf{X}_0^{2T}$, $k\in\mathbb{Z}^+$ of the initial distribution 
$\mu_{T,\theta}^0\in\mathcal{P}(\mathsf{X}_0^{2T})$ as in \eqref{eq:id_0} and transition kernel $K^{0}_{T,\theta}$ as described in Algorithm \ref{algo:ccpf} up to time $M=\max(\tau_{0},k^\star)$ (for $k^\star\in\{2,3,\dots\}$ and $m^\star>k^\star$). Then, we set
\begin{equation}
\begin{split}
\Psi_{T,\theta}^0 &:= \dfrac{1}{m^\star-k^\star+1}\sum_{k=k^\star}^{m^\star}\lambda_{T,\theta}^0(x_*,X_{[\Delta_0:T]}^{0,k}) + \\
&\sum_{k=k^\star+1}^{\tau^0-1}\dfrac{\min(m^\star-k^\star+1,k-k^\star)}{m^\star-k^\star+1}\{\lambda_{T,\theta}^0(x_*,X_{[\Delta_0:T]}^{0,k})-\lambda_{T,\theta}^0(x_*,\mathring{X}_{[\Delta_0:T]}^{0,k})\}.  \label{eq:psi_1}
\end{split}
\end{equation}
The quantity $\Psi_{T,\theta}^0$ corresponds to the Rhee-Glynn estimator, as described in Section \ref{sec:rhee-glynn_estimator}, of the gradient of the log-likelihood.

In contrast, $\Psi_{T,\theta}^l$, for $l>0$, is based on a Markov chain $(V_{[\Delta_{l}:T]}^{(l,0)},V_{[\Delta_{l-1}:T]}^{(l-1,0)}),\dots$, $(V_{[\Delta_{l}:T]}^{(l,k)},V_{[\Delta_{l-1}:T]}^{(l-1,k)})\in \mathsf{X}_l^{2T}\times \mathsf{X}_{l-1}^{2T}$, $k\in\mathbb{Z}^+$ of the initial distribution $\mu_{T,\theta}^{l,l-1}\in\mathcal{P}(\mathsf{X}_l^{2T}\times \mathsf{X}_{l-1}^{2T})$ as in \eqref{eq:id_l_l1} and transition kernel $\check{K}_{\theta}^{l,l-1}$ as described in Algorithm \ref{algo:cccpf} up to time $M=\max(\tau^\star, k^\star)$ (for $k^\star\in\{2,3,\dots\}$ and $m^\star>k^\star$), given $\tau^\star=\max(\tau^{l},\tau^{l-1})$ and $\tau^l = \inf\{k\geq 1:X_{[\Delta_l:T]}^{(l,k)}=\mathring{X}_{[\Delta_l:T]}^{(l,k)}\}$. Then, we set
\begin{eqnarray}
\Psi_{T,\theta}^l & := &
\dfrac{1}{m^\star-k^\star+1}\sum_{k=k^\star}^{m^\star}\lambda_{T,\theta}^l(x_*,X_{[\Delta_l:T]}^{l,k}) -
\lambda_{T,\theta}^{l-1}(x_*,X_{[\Delta_{l-1}:T]}^{l-1,k}) \nonumber \\ 
&+& \sum_{k=k^\star+1}^{\tau^\star-1}\dfrac{\min(m^\star-k^\star+1,k-k^\star)}{m^\star-k^\star+1}\Big\{\Big(\lambda_{T,\theta}^l(x_*,X_{[\Delta_{l}:T]}^{l,k})-\lambda_{T,\theta}^l(x_*,\mathring{X}_{[\Delta_{l}:T]}^{l,k})\Big) \nonumber \\
&-& \Big(\lambda_{T,\theta}^{l-1}(x_*,X_{[\Delta_{l-1}:T]}^{l-1,k})-\lambda_{T,\theta}^{l-1}(x_*,\mathring{X}_{[\Delta_{l-1}:T]}^{l-1,k})\Big)
\Big\}.\label{eq:psi_2}
\end{eqnarray}
Thus, based on \eqref{eq:psi_1}-\eqref{eq:psi_2}, when $L$ is sampled from $p^{\star}$, our estimator is as follows:
\begin{equation}\label{eq:estimate_single}
\frac{\Psi_{T,\theta}^{L}}{p^{\star}(L)}.
\end{equation}
The main task is now to verify that \eqref{eq:psi_1}-\eqref{eq:psi_2} have properties 2 and 3 listed in Section \ref{sec:main_idea}.

\begin{rem}
In practice, one can use an average estimator. Let $L^1,\dots,L^M$ be independent and identically distributed (i.i.d.) samples from $p^{\star}$. Then, independently, for each $L^i$, $i\in\{1,\dots,M\}$, obtain $\Psi_{T,\theta}^{L_i,i}$. One can then use
\begin{equation}\label{eq:estimate_avg}
\frac{1}{M}\sum_{i=1}^M\frac{\Psi_{T,\theta}^{L_i,i}}{p^{\star}(L^i)}
\end{equation}
to estimate \eqref{eq:gll}. Another alternative is the coupled sum estimator in \cite{rhee}: set $P^{\star}(l) = \sum_{p=l}^{\infty}p^{\star}(p)>0$; then, one samples
$L$ from $p^{\star}$ and constructs the estimator
\begin{equation}\label{eq:estimate_coup}
\sum_{l=0}^L \frac{\Psi_{T,\theta}^{l}}{P^{\star}(l)}.
\end{equation}
\end{rem}

\subsection{Sketch of Proof of Unbiasedness}\label{sec:theory}

To verify that \eqref{eq:psi_1}-\eqref{eq:psi_2} have properties 2 and 3 listed in Section \ref{sec:main_idea}, one can follow the blueprints in \cite{ub_mcmc,ub_grad}.
The approach in this paper is simply a modification of the methodology in \cite{ub_grad}: thus, although the strategy of the proof may be the same, the process that is
considered in this paper is more challenging, as it is necessary to average over the uncertainty in the data.
The stopping time was generally dealt with in \cite{ub_grad}; therefore so the main task is to demonstrate that the expectation of summands in the estimates \eqref{eq:psi_1}-\eqref{eq:psi_2} is as small as a function of $l$. The latter task requires one to consider the intricate properties of the C-CCPF and CCPF on an iteration-by-iteration basis, which in turn relies on the complex coupled particle filters that underly the iterations. Nonetheless, this has been achieved for a simpler process in \cite{ub_grad}, and we believe that a similar method can be used.

To select the distribution $p^{\star}$, we believe that one can use the recommendations in \cite{rhee} under Euler discretization when $\sigma$ is either constant or non-constant.
In either case, as in \cite{ub_grad}, this leads to an estimator that is unbiased with finite variance but with infinite expected cost. Nonetheless, with high probability, the estimator has finite cost. 

\section{Simulations}\label{sec:simos}

We discuss two choices of underlying distribution $p^{\star}(l)$ in the construction of unbiased estimator \eqref{eq:estimate_coup}. We consider geometric distribution $\mathcal{G}(p)$ with success rate $p=0.6$ and $p^{\star}(l)\propto \Delta^{1/2}_l (l+1)(\log_2(2+l))^2$ as suggested in \cite{ub_grad,rhee}. 
Then we compare estimator \eqref{eq:estimate_coup} built over these two underlying distributions with Rhee-Glynn estimator \eqref{eq:psi_1} for an increased number of particles $N$. We first compare the mean square error (MSE) satisfied by the estimators, and then visualize how this analysis reflects in a stochastic gradient descent (SGD) procedure to recover unknown parameters.

The workstation has 62.9 GiB of memory and Intel Xeon processor with forty CPUs ES-2680 with 2.80 GHz; the operating system is Ubuntu 18.04.5 LTS. The numerical test are programmed in \textit{python 3.8.5} and the wall clock times are measured using library \textit{timeit}.

\subsection{Model Settings} 
  
The diffusion process we consider is as follows
\begin{eqnarray*}   
\mathrm{d}Y_{t}&=&h_{\theta}(X_{t})\mathrm{d}t+\mathrm{d}B _{t},   \\
\mathrm{d}X_{t}&=&b_{\theta}(X_{t})\mathrm{d}t+\sigma(X_t)\mathrm{d}W _{t}   
\end{eqnarray*}
with $0\leq t \leq T$ and starting points $X_{0}=x_{\star}$ and $Y_{0}=y_{\star}$.
Here $\lbrace W_{t}\rbrace_{t\in [0,T]}$ and $\lbrace B_{t}\rbrace_{t\in [0,T]}$  are independent Brownian motions, and the final time is $T=50$. 

\subparagraph{Ornstein-Uhlenbeck (OU)} 
\begin{eqnarray*}
\mathrm{d}Y_{t}&=&\theta_{1} (\mu_{1}-X_t)\mathrm{d}t+\mathrm{d}B_{t},\\
\mathrm{d}X_{t}&=&-\theta_{2} X_t\mathrm{d}t+\sigma \mathrm{d}W_{t},  
\end{eqnarray*}
with $0\leq t \leq T$ and parameters $\theta_1=0.75$, $\theta_2=0.75$, $\mu_{1}=1$, and $\sigma=0.5$. The starting points $X_{0}$ and $Y_{0}$ are sampled independently from the normal distribution $\mathcal{N}(0, 1.6\cdot 10^{-3})$.

\subparagraph{Geometric Brownian Motion (GBM)} 
\begin{eqnarray*}
\mathrm{d}Y_{t}&=&\theta_{1} (\mu_{1}-\log(X_t))\mathrm{d}t+\mathrm{d}B_{t}, \\
\mathrm{d}X_{t}&=&\theta_{2} X_t\mathrm{d}t+\sigma X_t \mathrm{d}W_{t},  
\end{eqnarray*}
with $0\leq t \leq T$ and parameters $\theta_1=0.75$, $\theta_2=0.05$, $\mu_{1}=1$, and $\sigma=0.05$. The starting point $X_{0}$ is sampled from the distribution $\mathcal{N}(5,1.6\cdot 10^{-3})$ while $Y_0$ is sampled from $\mathcal{N}(0,1.6\cdot 10^{-3})$. 

\subparagraph{Lorenz Model (LM)} 
\begin{eqnarray*}
\mathrm{d}X_{1,t}&=&-S(X_{1,t}-1)\mathrm{d}t+\mathrm{d}W_{1,t},  \\
\mathrm{d}X_{2,t}&=&(X_{1,t}-BX_{2,t})\mathrm{d}t+\mathrm{d}W_{2,t},   \\
\mathrm{d}Y_{1,t}&=&k X_{1,t}\mathrm{d}t+\mathrm{d}W_{3,t},   \\
\mathrm{d}Y_{2,t}&=&k X_{2,t}\mathrm{d}t+\mathrm{d}W_{4,t},  
\end{eqnarray*}
with $0\leq t \leq T$ and parameters $S=10$, $B=8/3$, and $k=2$. The starting points $X_{1,0}$, $X_{2,0}$ and $Y_{1,0}$, $Y_{2,0}$ are sampled independently from the distribution $\mathcal{N}(0,1.6\cdot 10^{-3})$. We define $\lbrace W_{i,t}\rbrace_{i=1}^{4}$ as an independent one-dimensional Wiener process.

\subsection{Algorithm Settings}

Level $l$ corrisponds to discretization $\Delta_l=2^{-(l+3)}$. In Algorithms \ref{algo:ccpf} and \ref{algo:cccpf}, we perform the resampling step when the effective sample size (ESS) is lower than $N/4$. Given iteration $k$ and level $l$ in Algorithm \ref{algo:ccpf}, the ESS is defined as 
\begin{eqnarray}
ESS=\Bigg(\sum_{j=1}^{N}w^{2}_j\Bigg)^{-1}
\end{eqnarray}
with 
\begin{eqnarray}
w_j=\frac{F_{k,\theta}^l(j,\mathbf{u}_{k,k}^{l,1:N})\wedge
F_{k,\theta}^l(j,\mathring{\mathbf{u}}_{k,k}^{l,1:N})}{\sum_{s=1}^N\{ F_{k,\theta}^l(s,\mathbf{u}_{k,k}^{l,1:N})\wedge
F_{k,\theta}^l(s,\mathring{\mathbf{u}}_{k,k}^{l,1:N})\}}.
\end{eqnarray}
In Algorithm \ref{algo:cccpf}, the ESS is defined over weights $F_{k,\theta}^{l-1}(\cdot,\mathbf{u}_{k,k}^{l-1,1:N})$, $F_{k,\theta}^{l-1}(\cdot,\mathring{\mathbf{u}}_{k,k}^{l-1,1:N})$.

We consider $\mathcal{S}=5$ i.i.d. time series $Y^s_{t}$ discretized on level $l^{\star}=11$. For each time series, we perform $\mathcal{R}=100$ i.i.d. evaluations of the estimators \eqref{eq:estimate_coup} and \eqref{eq:psi_1}. We consider $k^{\star}=2$ and $m^{\star}=4$, where the method of selecting these parameters is based on the analysis of hitting times $\tau$ as described in \cite{rhee}.

For OU and GBM, we consider $N=\{128,256,512,1024\}$, while for the LM model, we consider $N=\{362,512,724,1024\}$. Thus, to compute the MSE given $N$, we must estimate the variance of both estimators and the bias of the Rhee-Glynn estimator \eqref{eq:psi_1} since estimator \eqref{eq:estimate_coup} is unbiased. 

To compute variance $V_l$ of terms $\Psi_{T,\theta}^{l}$ for $l>0$, for each time series, we estimate sample variance $V_{l,s}$ for $N=1024$ over $100$ repeats for each time series. Then we average over $\mathcal{S}$ quantities to obtain $V_{l}$. Similarly, to assess the variance of estimator \eqref{eq:estimate_coup}, for each time series, we compute the sample variance $V_{N,s}$ over $100$ repeats for each time series and we average over $\mathcal{S}$ quantities to obtain $V_{N}$.
The bias of the Rhee-Glynn estimator \eqref{eq:psi_1} is computed by evaluating (for each time series) the $95\%$ percentile of $100$ realizations of $\Psi_{T,\theta}^1$, and then averaging over $\mathcal{S}$ quantities.

We wish to recover parameter $\theta_1$ in the OU and GBM cases, and parameter $k$ in LM by SGD specified in Algorithm \ref{alg:SGD} with $N=2^{10}$. We present the hyper-parameters for each model in Table \ref{table:SGD}. For each time series, we perform SGD $10$ times with different initializations.
Coherently with observed data $Y^{s}_t$ discretized on level $l^{\star}=11$, the empirical distribution $p^{\star}(l)\propto \Delta^{1/2}_l (l+1)(\log_2(2+l))^2$ is normalized over levels $l\in\{0,\dots,8\}$. 
 
 \begin{algorithm}[H]
\begin{enumerate}
\item Initialization $\theta$ given distribution $\mu(\cdot)$, learning step $\alpha$, $i=0$, $k=0$
\item Compute $\xi_0=\log(\theta)$
\item While $k\leq 1000$ and $i<10$:
\begin{itemize}
\item Compute $\varphi_k$ by \eqref{eq:psi_1} or \eqref{eq:estimate_coup}
\item Update $\xi_{k+1}=\xi_k+\alpha\varphi_k\exp(\xi_k)$
\item If $|\exp(\xi_{k+1})-\exp(\xi_{k})|<\beta$, then $i=i+1$, otherwise $i=0$    
\item If $(k \mod 50)=0$ then $\alpha=\alpha/2$
\item Increase $k=k+1$ 
\end{itemize}
\item return $\theta=\exp(\xi_k)$
\end{enumerate}
\caption{Stochastic Gradient Descent (SGD)}
\label{alg:SGD}
\end{algorithm}
 
\subsection{Results}

\begin{figure}[H]%
\subfloat[\pmb{Ornstein-Uhlenbeck (OU)}.]{\label{fig:OU_var_conv}
\includegraphics[width=.5\textwidth]{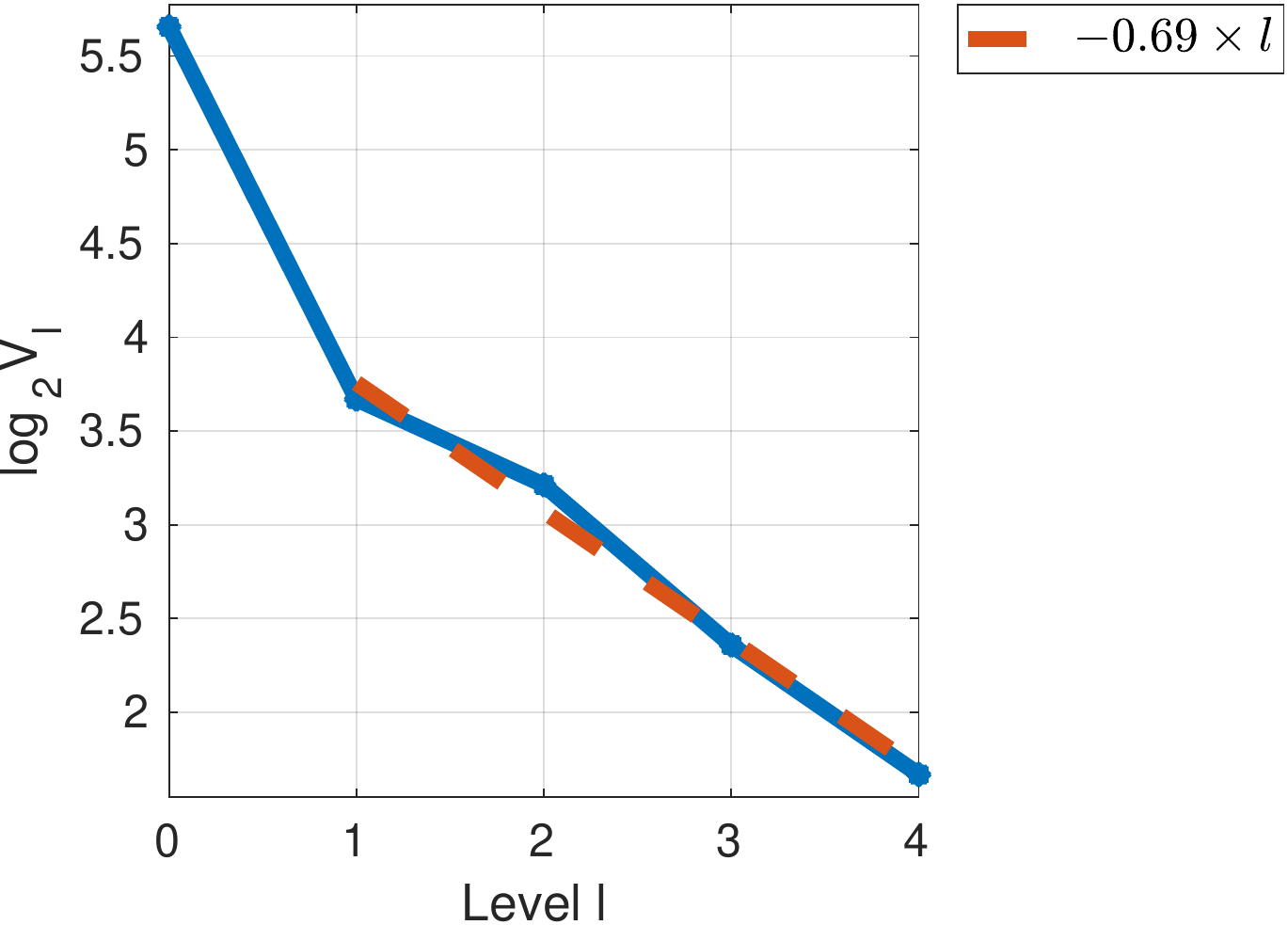}
}
\subfloat[\pmb{Geometric Brownian motion (GBM).}]{\label{fig:GBM_var_conv}
\includegraphics[width=.475\textwidth]{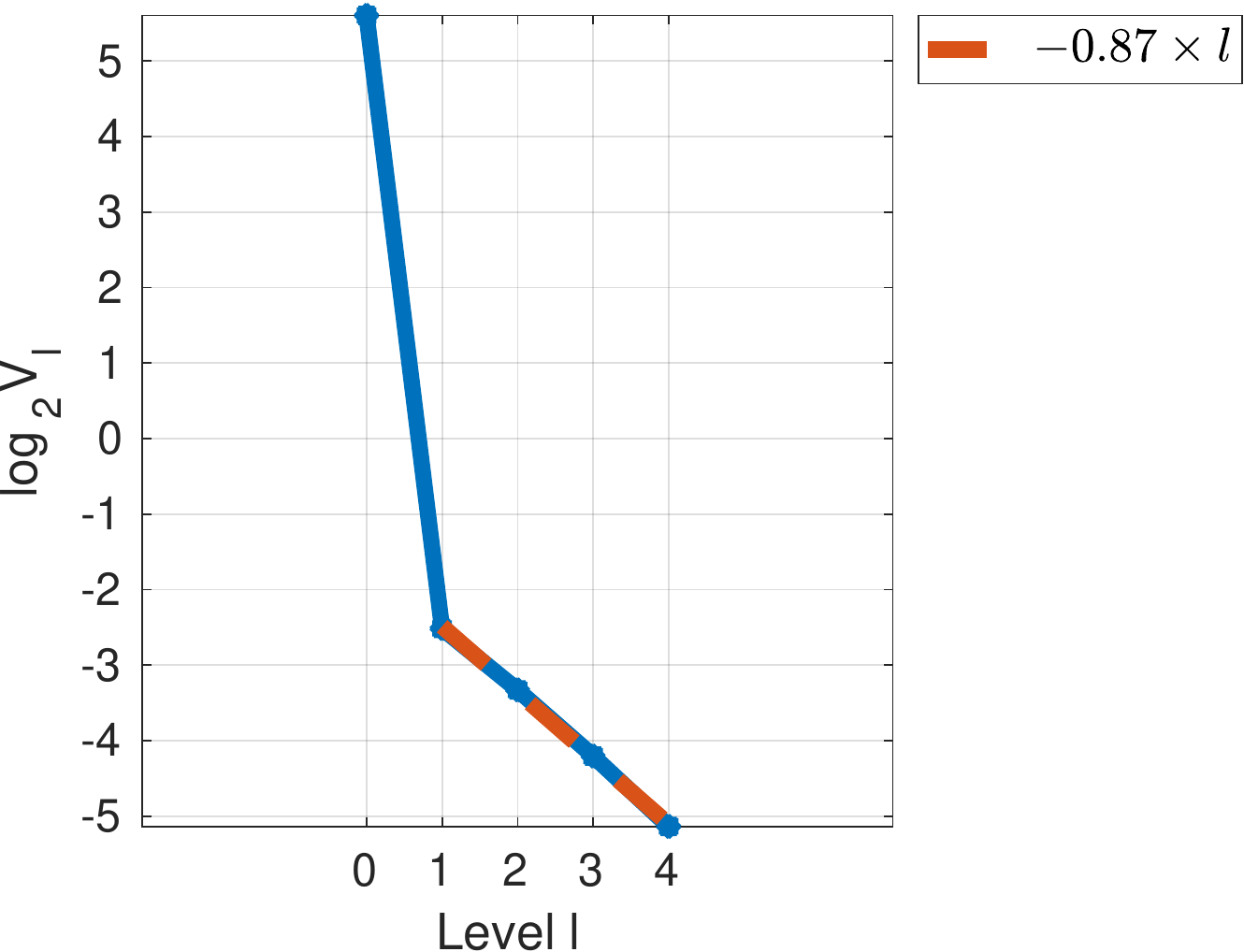}
} 

\vspace{0.3cm}

\centering
\subfloat[\pmb{Lorenz model (LM)}.]{\label{fig:LM_var_conv}
\includegraphics[width=.5\textwidth]{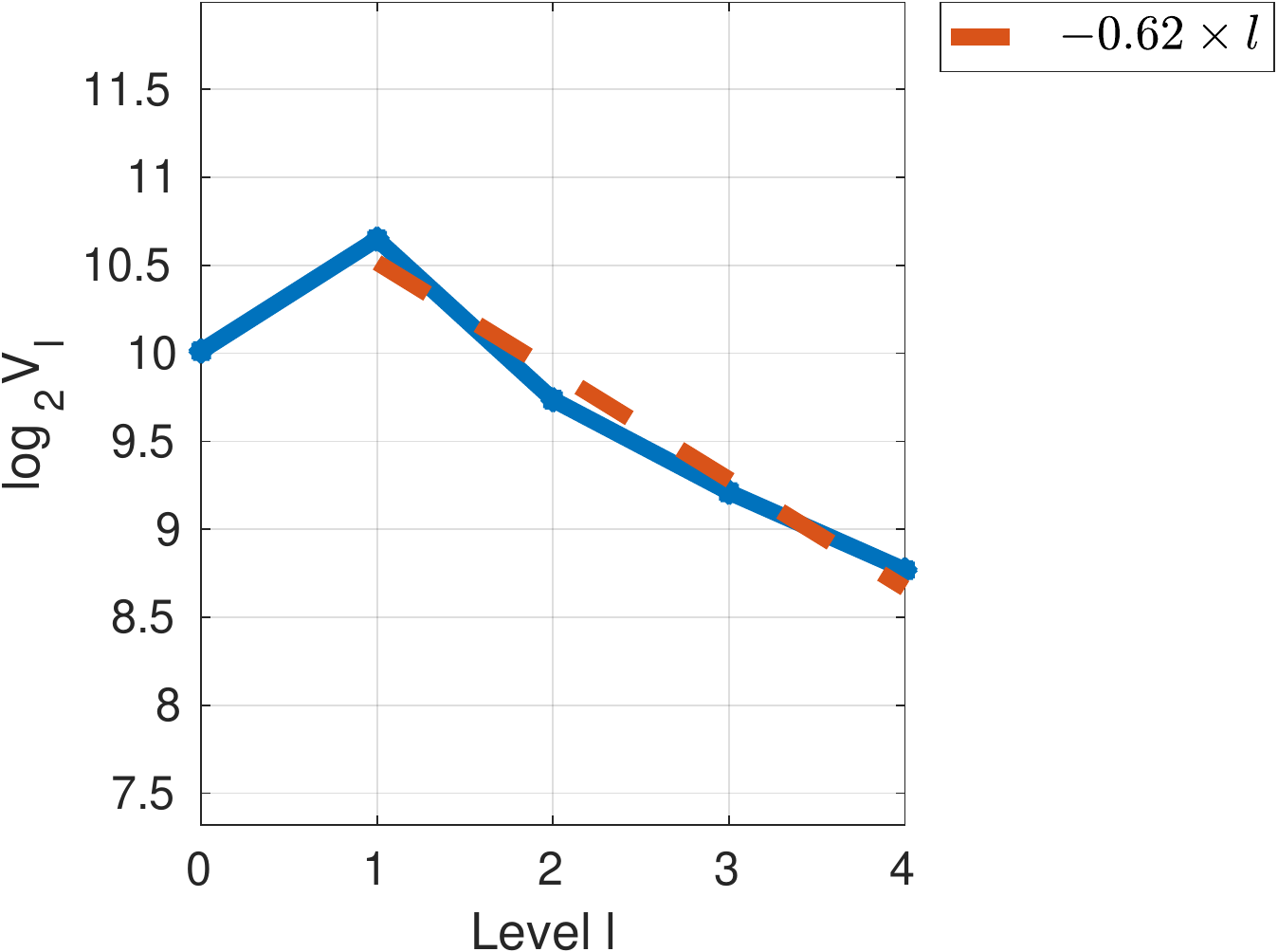}
}
\caption{Variance $\Psi_{T,\theta}^{l}$ for $l=\{0,\dots,4\}$.}
\label{fig:var_conv}
\end{figure}

In Figures \ref{fig:OU_var_conv}, \ref{fig:GBM_var_conv}, and \ref{fig:LM_var_conv}, we display the variance convergence of $\Psi_{T,\theta}^l$ for $l\geq 0$, $V_l$, respectively for the OU and GBM cases and LM. The convergence rates are lower than the Euler-Maruyama numerical scheme alone. The reason is the resampling procedure as described in Algorithm \ref{alg:thorrison} implemented when ESS is lower than $N/4$, indeed, resampling is applied to avoid ensemble collapse, but ruins the variance convergence rate (see e.g.~\cite{mlpf}).

\begin{figure}[H]
\centering
\includegraphics[width=.65\textwidth]{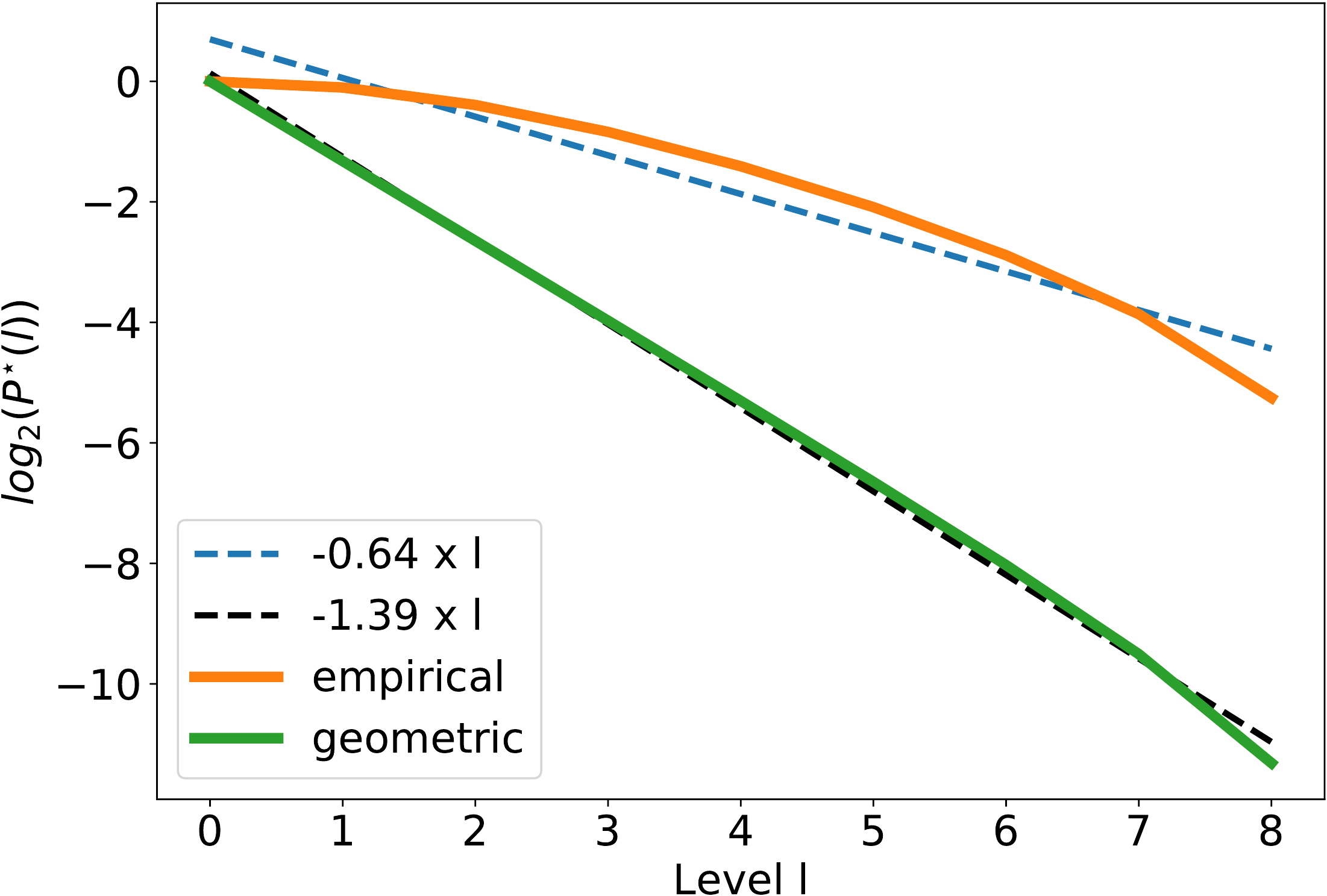}
\caption{Survival function $P^{\star}(l)$ for geometric distribution with success rate $p=0.6$ and empirical distribution $p^{\star}(l) \propto \Delta^{1/2}_l(l+1)(\log_{2}(2+l))^{2}$.}\label{fig:under_distribution}
\end{figure}

\begin{figure}[H]
\centering
\subfloat[\pmb{Ornstein-Uhlenbeck (OU)}]{\label{fig:OU_Var}
\includegraphics[width=.425\textwidth,height=6cm]{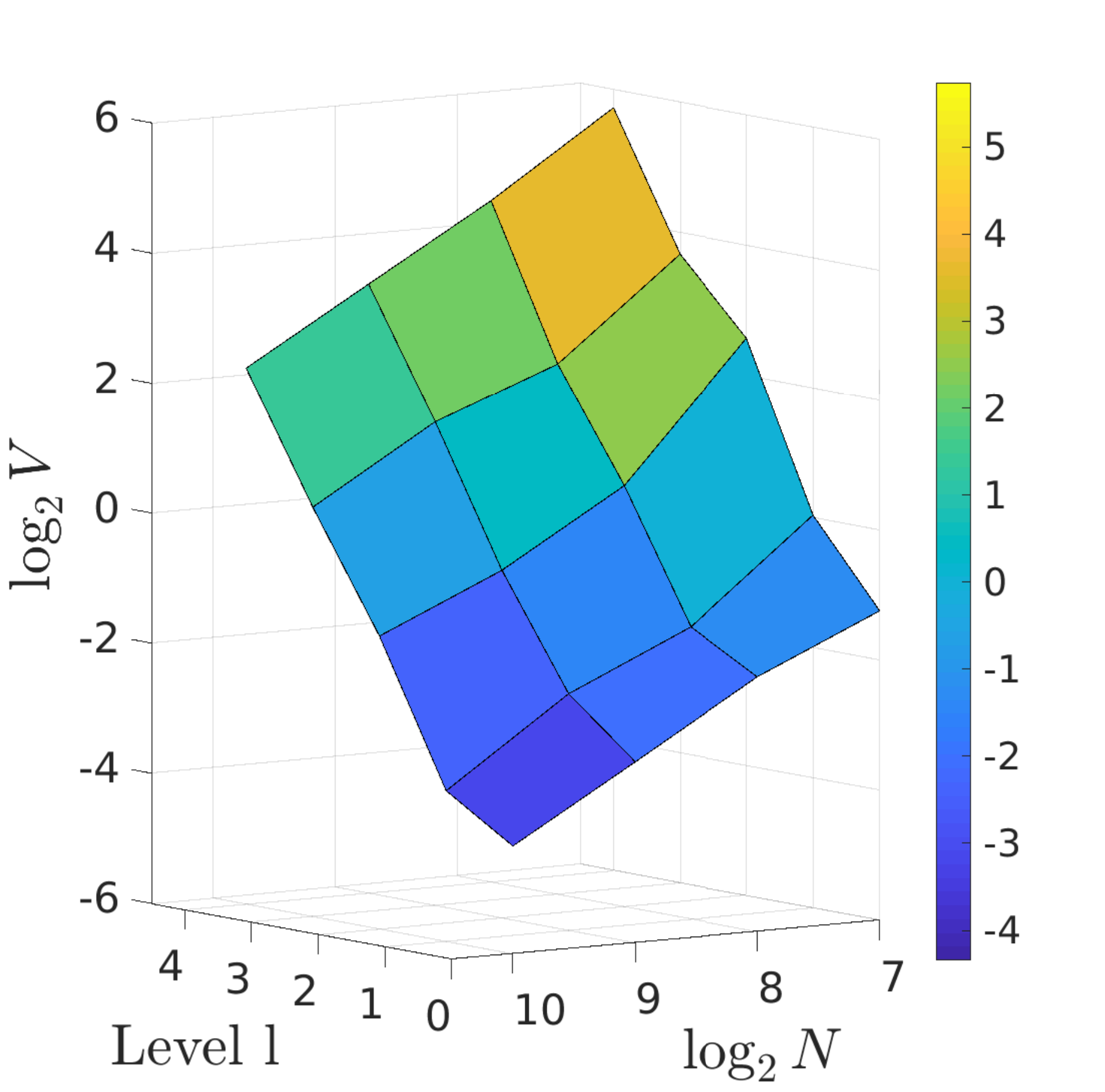}
\includegraphics[width=.425\textwidth,height=5.85cm]{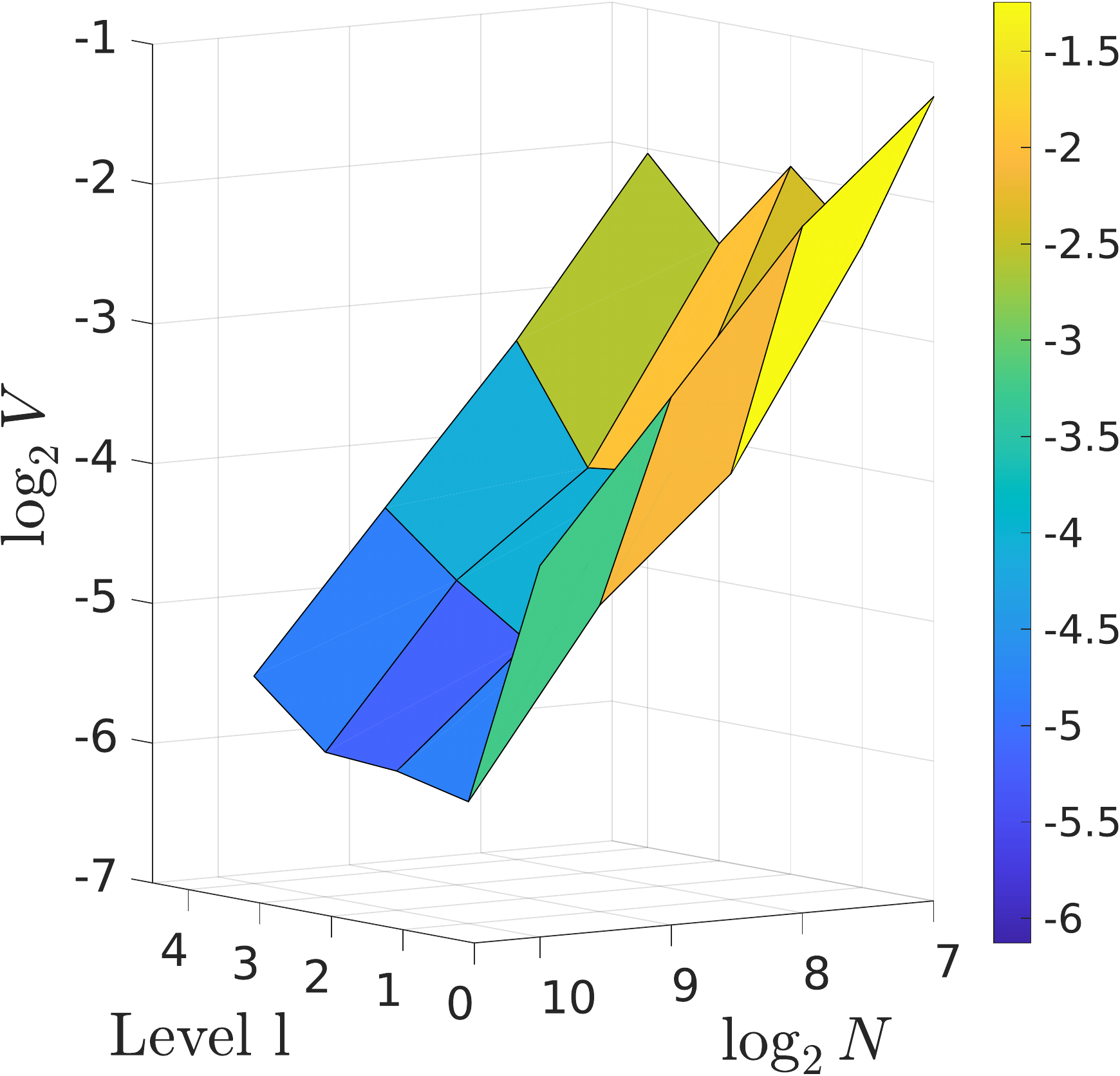}
}\quad
\subfloat[\pmb{Geometric Brownian motion (GBM).}]{\label{fig:GBM_Var}
\includegraphics[width=.425\textwidth,height=6cm]{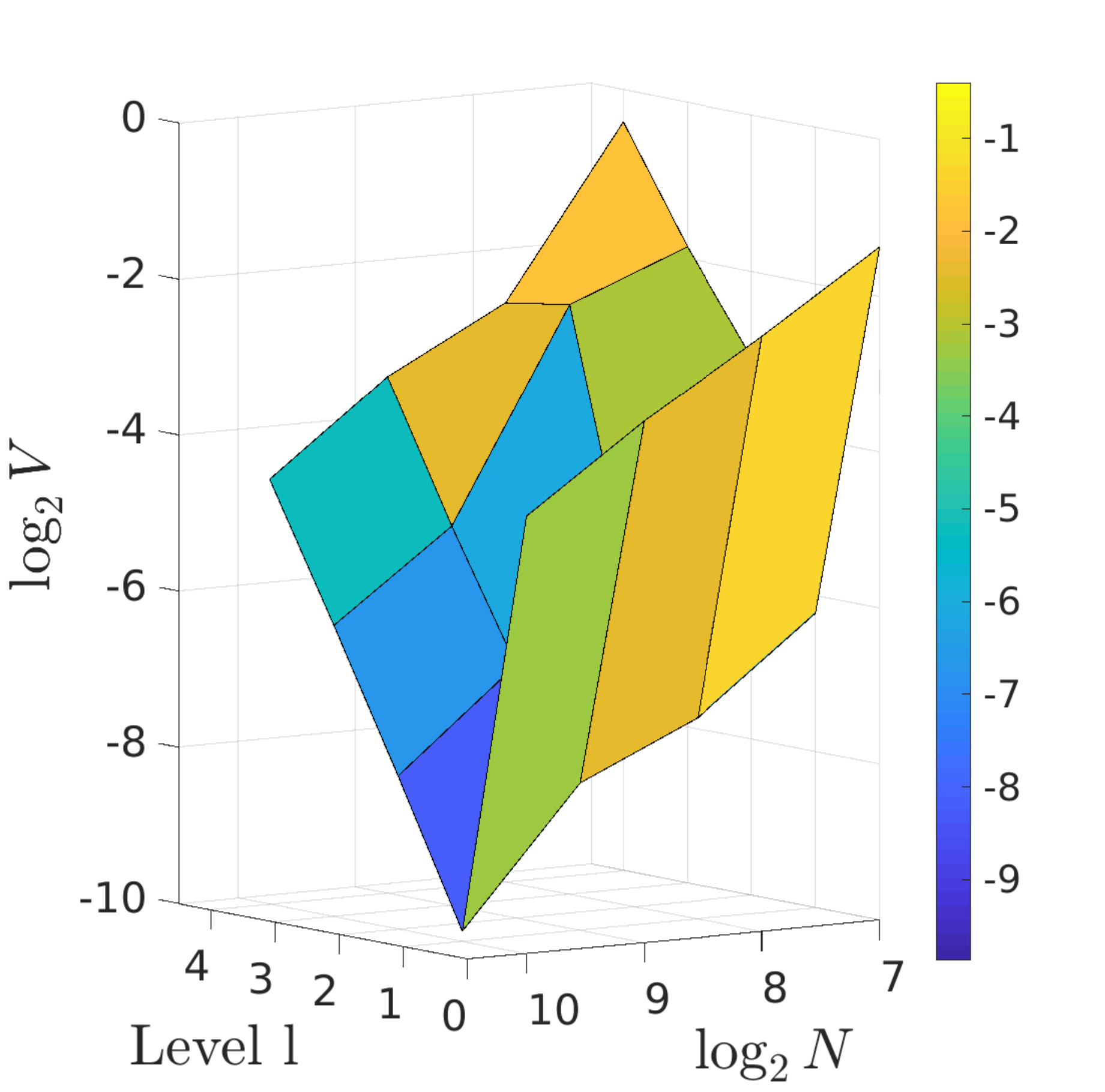}
\includegraphics[width=.425\textwidth,height=5.85cm]{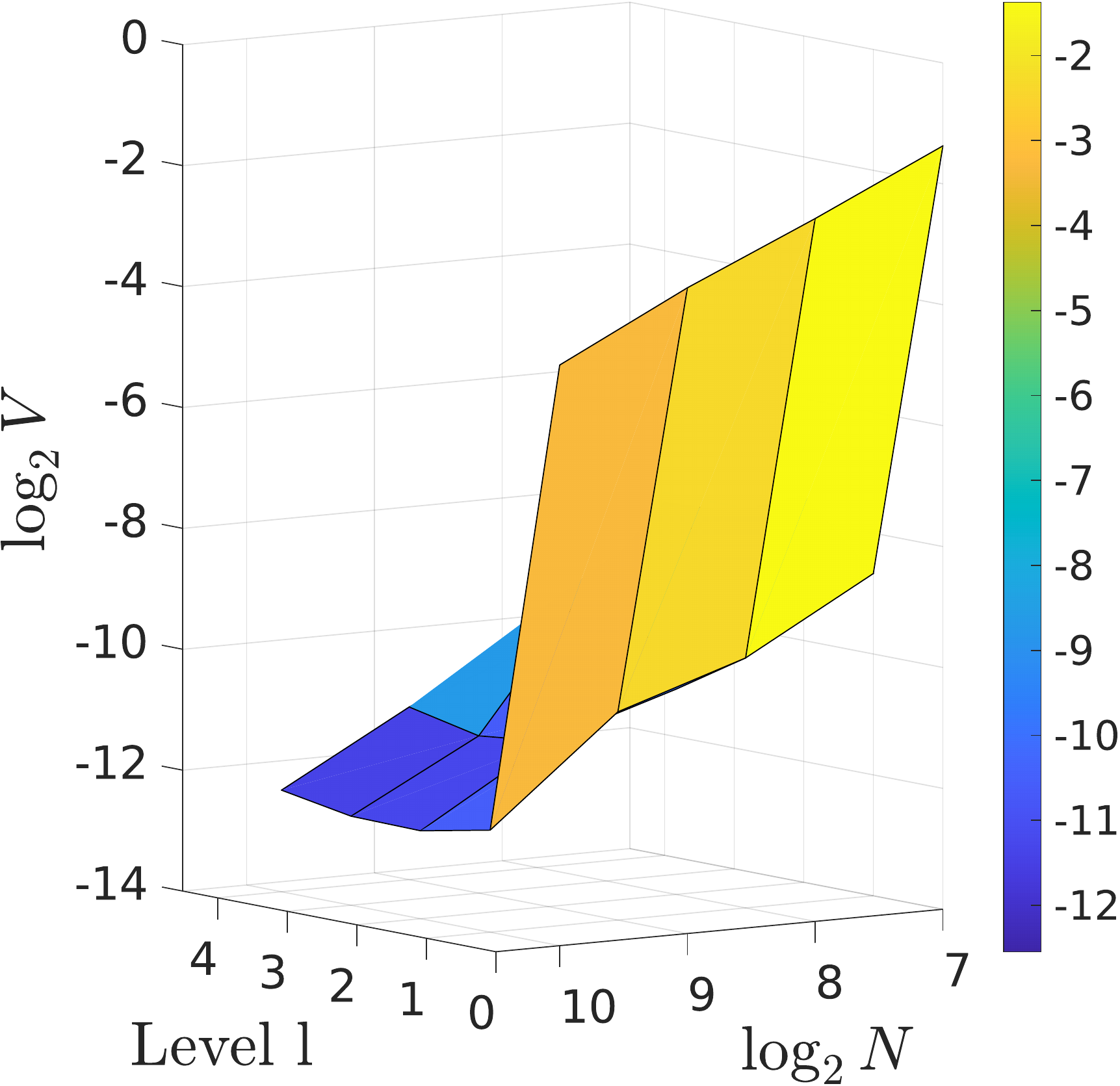}
}\quad
\subfloat[\pmb{Lorenz model (LM)}.]{\label{fig:LM_Var}
\includegraphics[width=.425\textwidth,height=6cm]{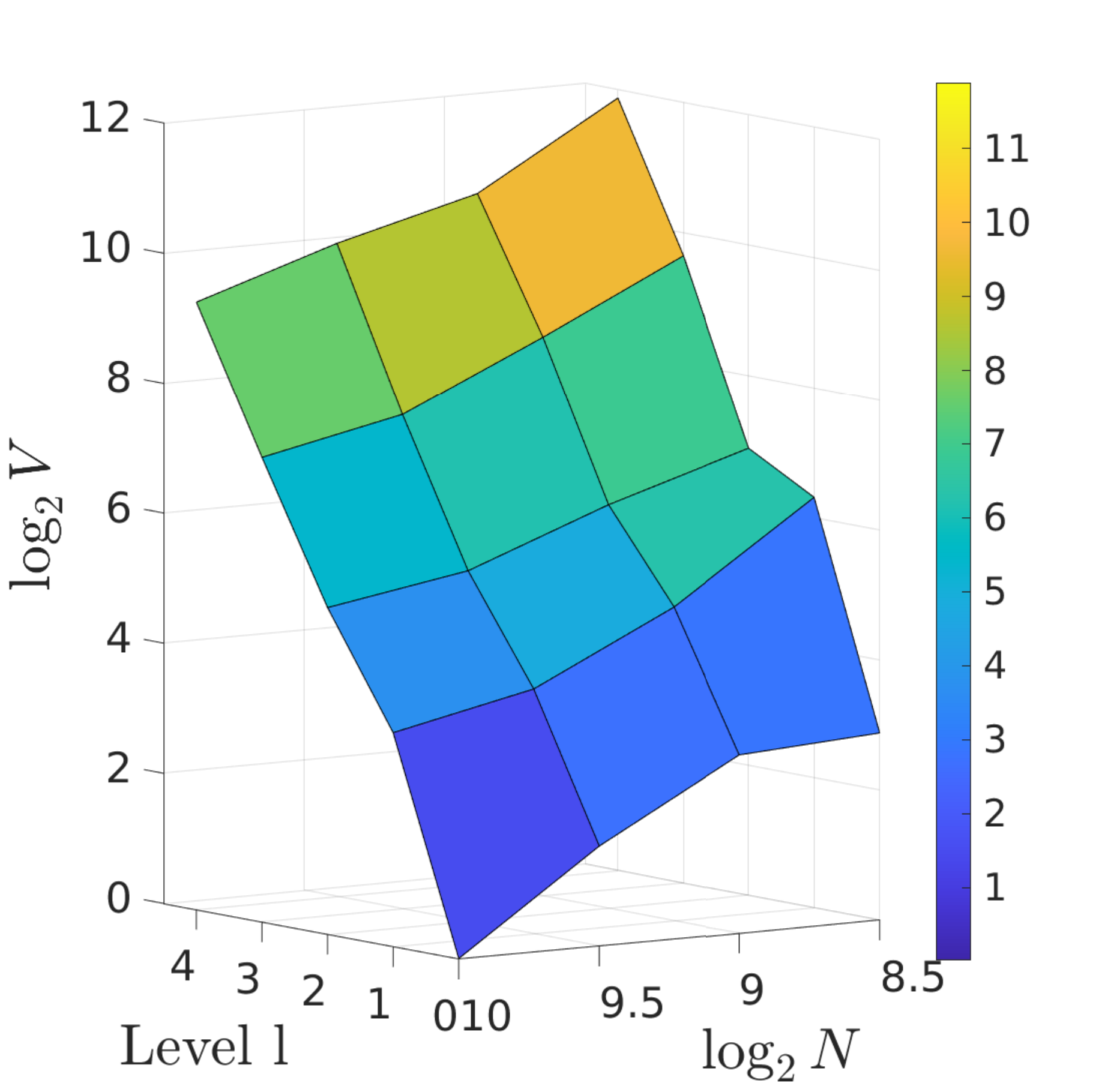}
\includegraphics[width=.425\textwidth,height=5.85cm]{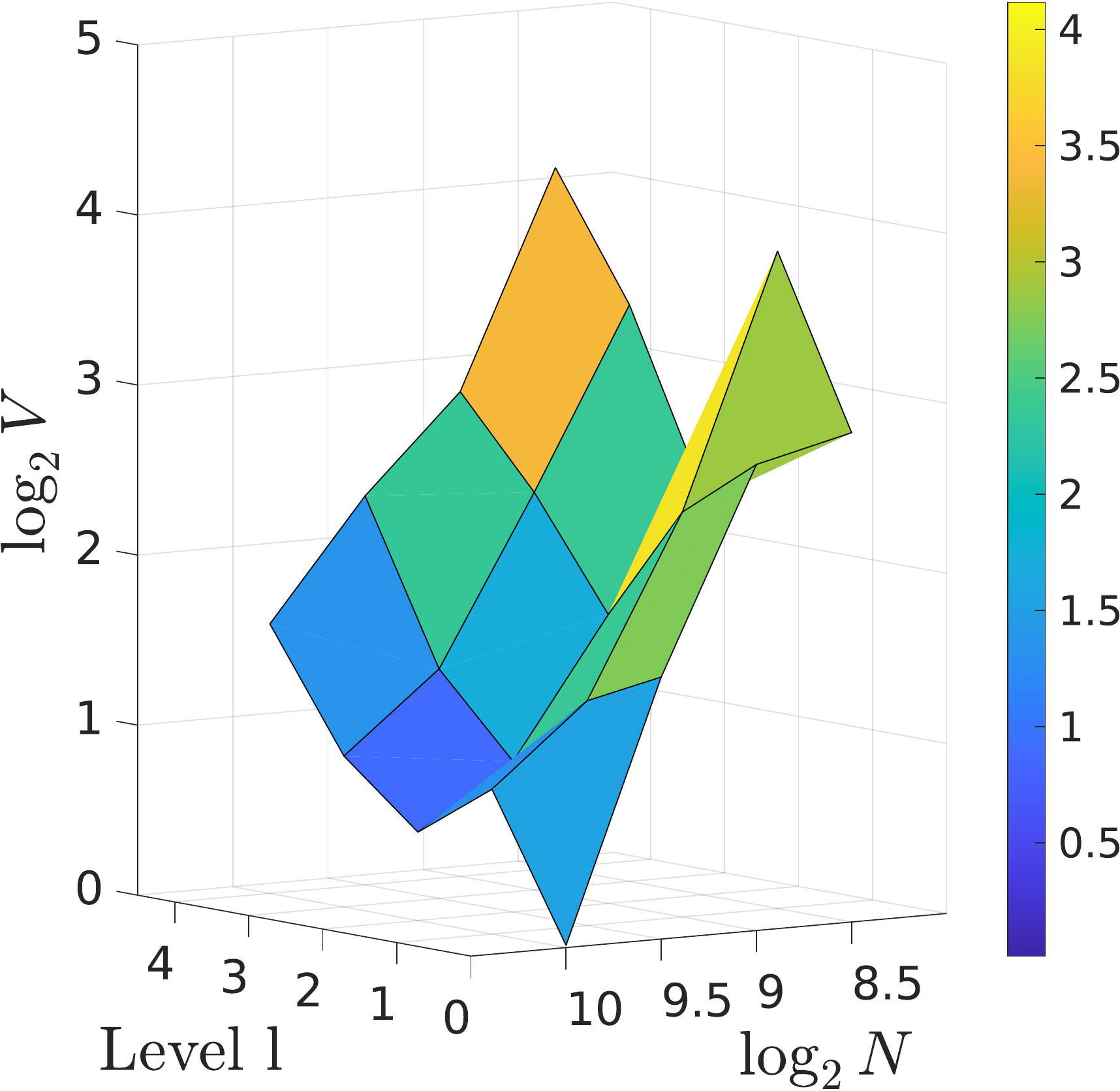}}%
\caption{Variance $\Psi_{T,\theta}^{l}/P^{\star}(l)$ for $l=\{0,\dots,4\}$ and $N=\{2^7,2^8,2^9,2^{10}\}$ for OU and GBM cases and $N=\{2^{8.5},2^9,2^{9.5},2^{10}\}$ for LM case. \pmb{Left figure:} Geometric underlying distribution, success rate $p=0.6$. \pmb{Right figure:} Empirical underlying distribution $p^{\star}(l) \propto \Delta^{1/2}_l(l+1)(\log_{2}(2+l))^{2}$.}
\label{fig:var}
\end{figure}

Choosing distribution $p^{\star}(l)$ over the level hierarchy is important to obtain a finite variance estimator \eqref{eq:estimate_avg}.  In Figures \ref{fig:OU_Var}, \ref{fig:GBM_Var}, and \ref{fig:LM_Var} we compute the variance of terms $\Psi_{T,\theta}^l/P^{\star}(l)$, for two choices of distribution $p^{\star}(l)$: a geometric distribution with success rate $p=0.6$ and $p^{\star}(l)\propto \Delta^{1/2}(l+1)(\log_2(2+l))^2$. For all numerical cases, variances explode moving on finest levels with $p^{\star}(l)$ distribution modeled as geometric distribution, while finite variance is achieved with empirical distribution. Such a behavior can be explained observing the survival function decay rate of the distributions in Figure \ref{fig:under_distribution}, and compare these rates with variance $V_l$ convergence rates of $\Psi_{T,\theta}^l$ in Figures \ref{fig:var_conv}. The survival function of the geometric distribution decreases with a rate of about $1.4$, while the one of the empirical distribution decays slower with a rate of about $0.64$.  While the geometric distribution rate is too high with respect to $V_l$ variance rates, empirical distribution survival function decay rate is lower than OU and GBM cases, and slighlty higher for the LM case, displaying an overall improvement of the variance of terms $\Psi_{T,\theta}^l/P^{\star}(l)$ and estimator \eqref{eq:estimate_coup}. The choice of the empirical distribution does not seem to be optimal for the LM case, but given the truncation of the level hierarchy for computational feasibility, a finite variance unbiased estimator is achieved anyways.

\begin{figure}[H]
\centering
\subfloat[\pmb{Ornstein-Uhlenbeck (OU)}.]{\label{fig:OU_MSE}
\includegraphics[width=.5\textwidth]{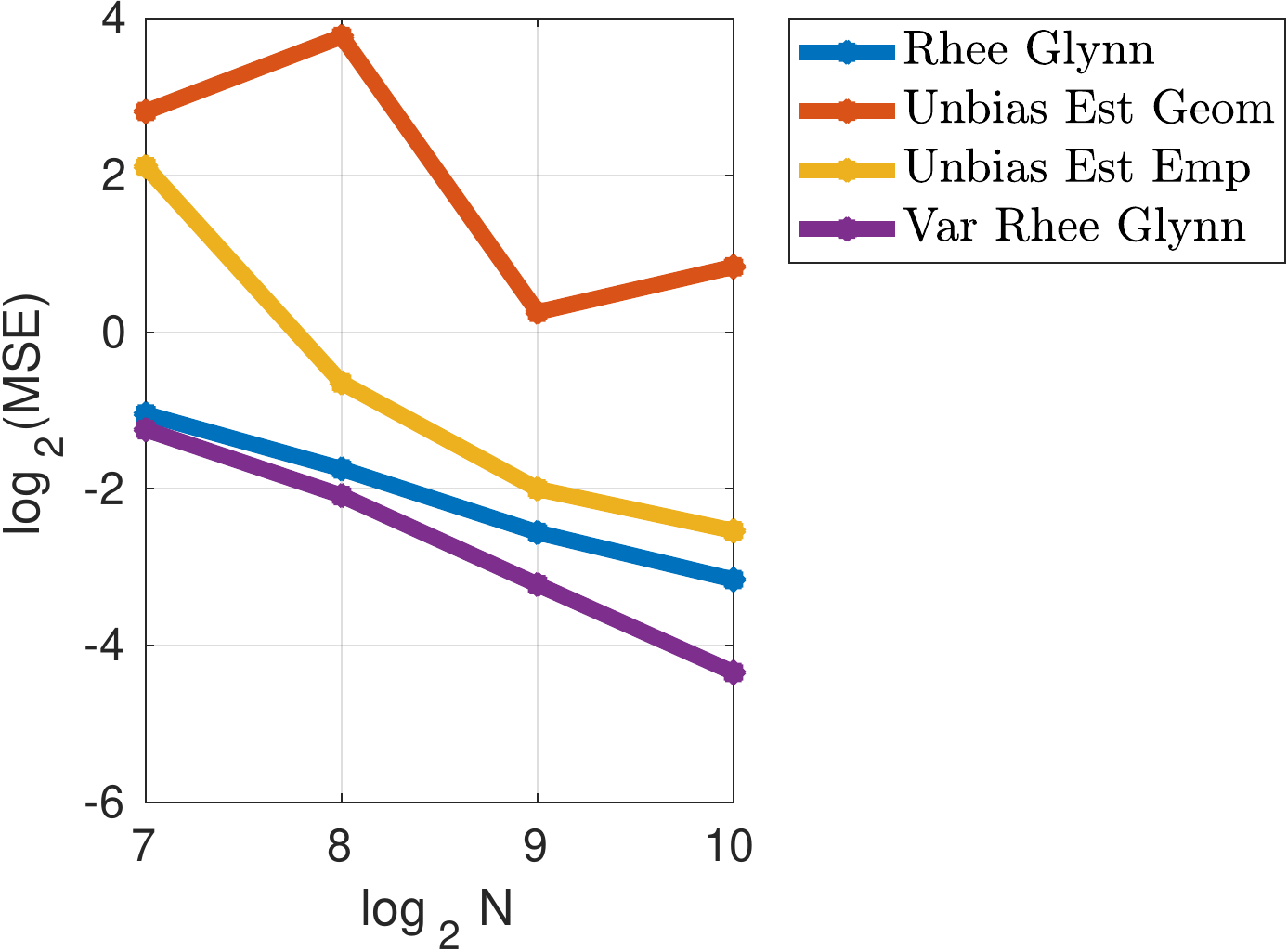}
\includegraphics[width=.5\textwidth]{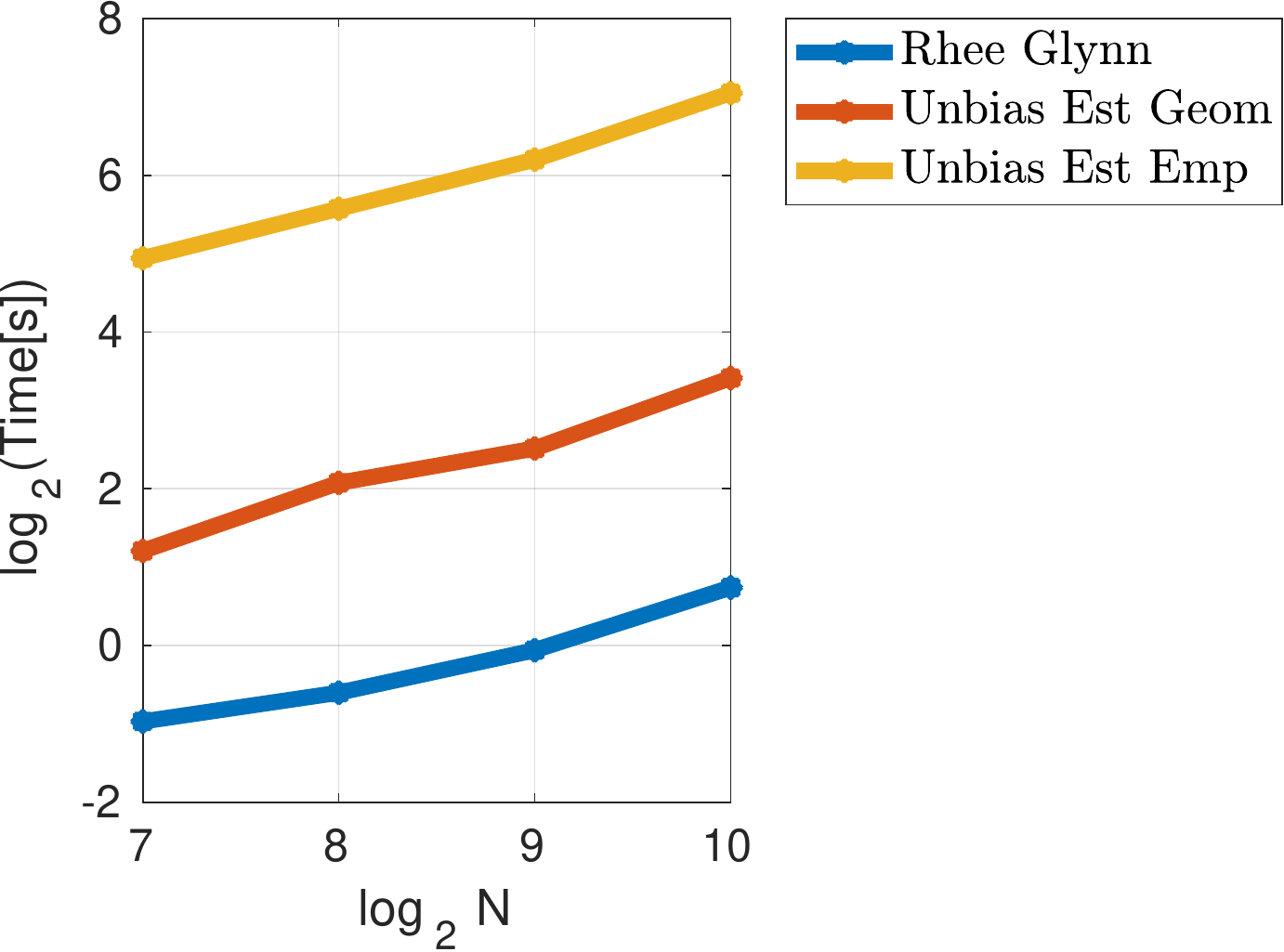}
}\qquad
\subfloat[\pmb{Geometric Brownian motion (GBM).}]{\label{fig:GBM_MSE}
\includegraphics[width=.5\textwidth]{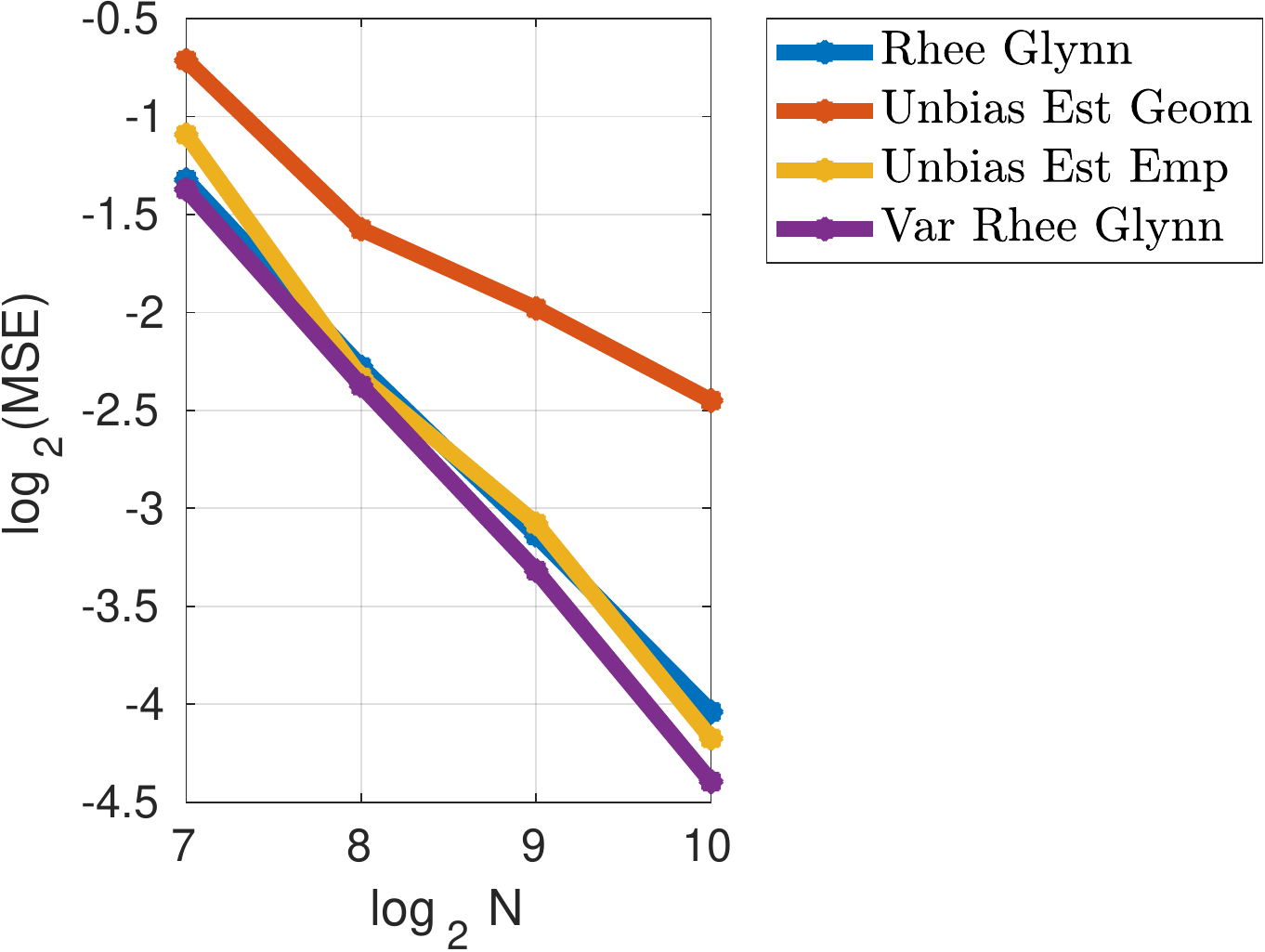}
\includegraphics[width=.5\textwidth]{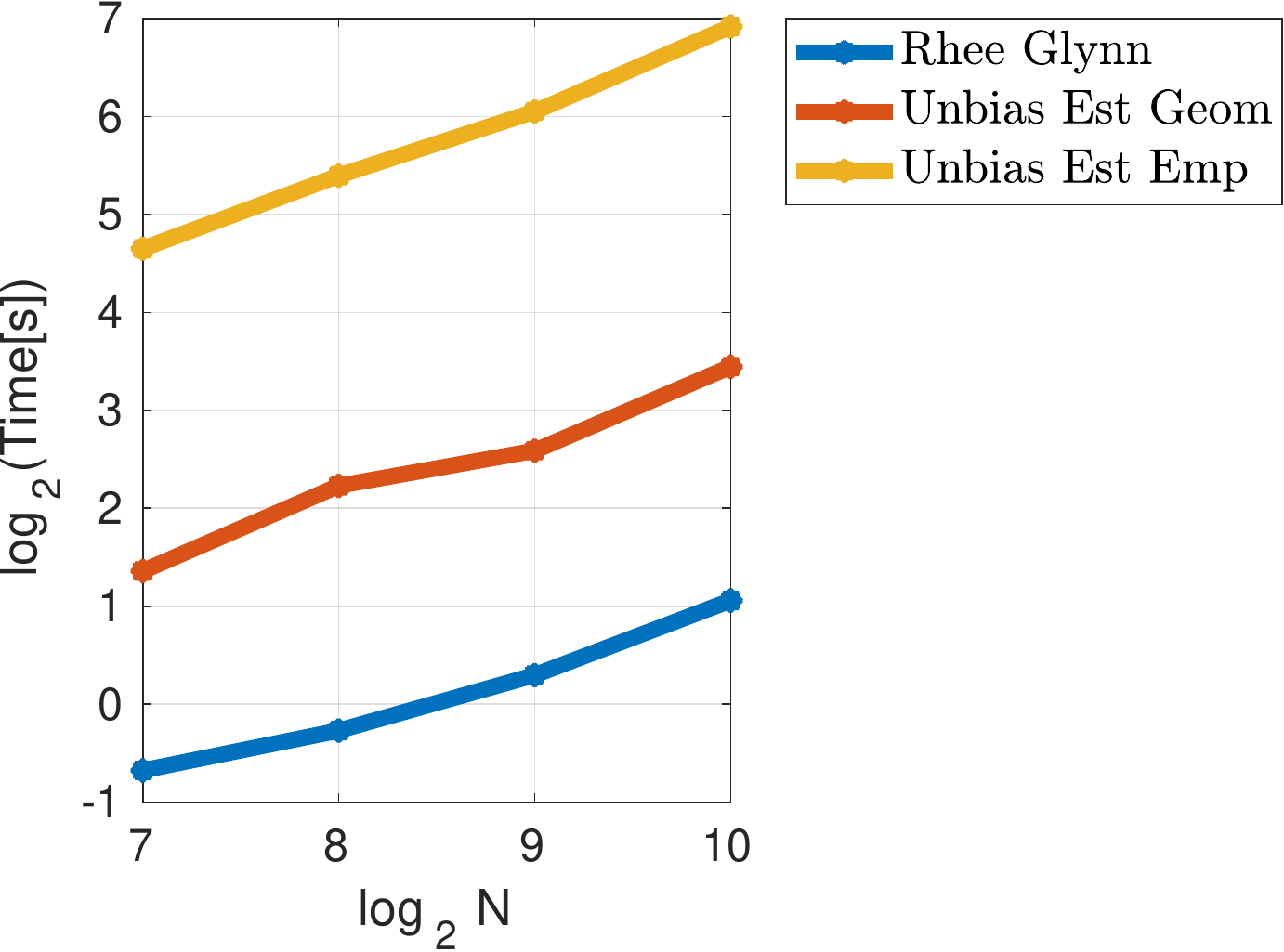}
}\qquad
\subfloat[\pmb{Lorenz model (LM)}.]{\label{fig:LM_MSE}
\includegraphics[width=.5\textwidth]{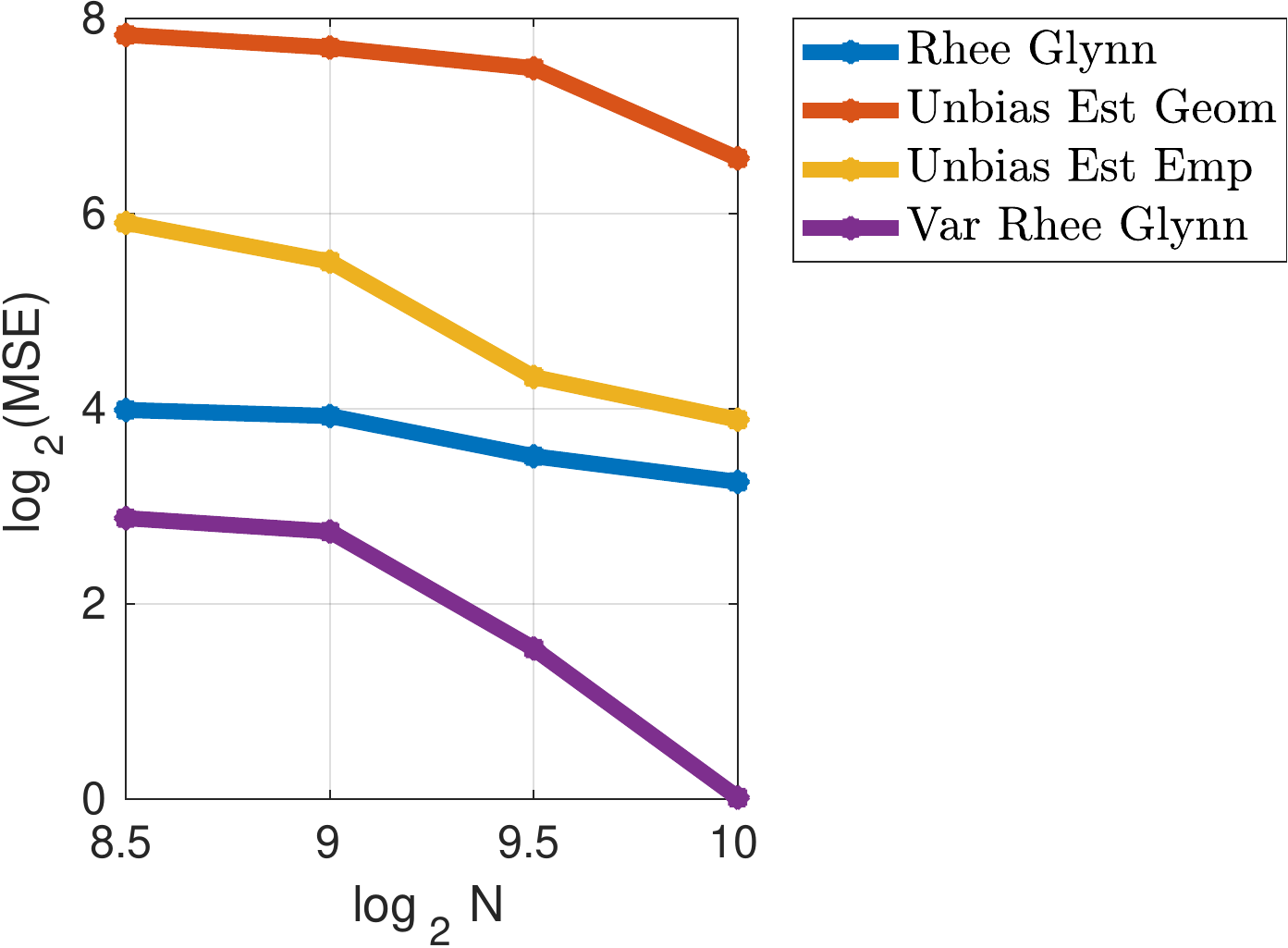}
\includegraphics[width=.5\textwidth]{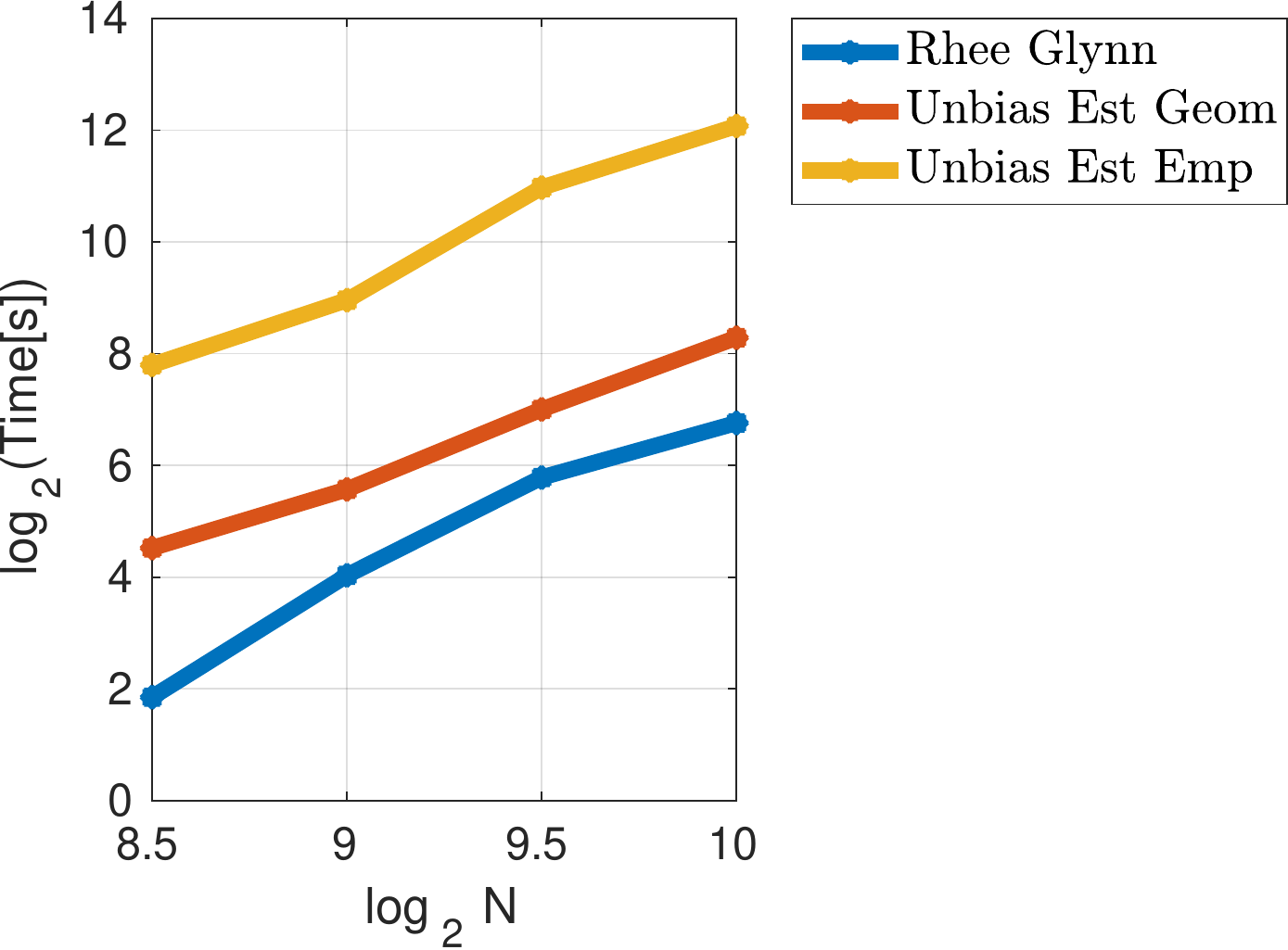}
}
\caption{\pmb{Left} \pmb{figure}: mean square error (MSE) achieved for fixed $N$. \pmb{Right} \pmb{figure}: computation time in seconds for fixed $N$.}
\label{fig:MSE}
\end{figure}

We display the MSE in Figures \ref{fig:OU_MSE}, \ref{fig:GBM_MSE}, and \ref{fig:LM_MSE} for a fixed number of particle ensemble $N$, respectively, for the OU and GBM cases and LM for estimators \eqref{eq:estimate_coup} and \eqref{eq:psi_1}. We can observe that the MSE achievable by unbiased estimator \eqref{eq:estimate_coup} with empirical distribution is lower than MSE that unbiased estimator can reach \eqref{eq:estimate_coup} with geometric distribution, consistently with previous variance analysis.  On the other side, we can observe as unbiased estimator \eqref{eq:estimate_coup} built over the empirical distribution is more computationally expensive than the one built over the geometric distribution. The reason is that, as can be deduced by the survival function displayed in \ref{fig:under_distribution}, geometric distribution has most of the mass on coarser levels, while empirical distribution weights the mass more uniformly on the level hierarchy. With the geometric distribution, mainly coarser and cheaper levels are sampled to build the unbiased estimator. In comparison, deeper and more expensive levels occur with higher probability when the empirical distribution is adopted. 

Unbiased estimators are compared with biased Rhee-Glynn estimator built on level $l=0$. We can observe that  Rhee-Glynn estimator, since evaluated on level $l=0$, results cheaper than unbiased estimators, especially with respect to the unbiased estimator built over the empirical distribution. The MSE solved by the Rhee-Glynn estimator is slightly lower than the one solved by the unbiased estimator \eqref{eq:estimate_coup} with empirical distribution for the OU and LM case and higher for the GBM case. The unbiased estimator with respect to Rhee-Glynn estimator has the advantage that bias is negligible since probability distributions have mass on levels up to frequency close to observed data $Y^{s}_t$. 

\begin{figure}[H]
\centering
\subfloat[\pmb{Ornstein-Uhlenbeck (OU)}.]{\label{fig:OU_sgd}
\includegraphics[width=.425\textwidth]{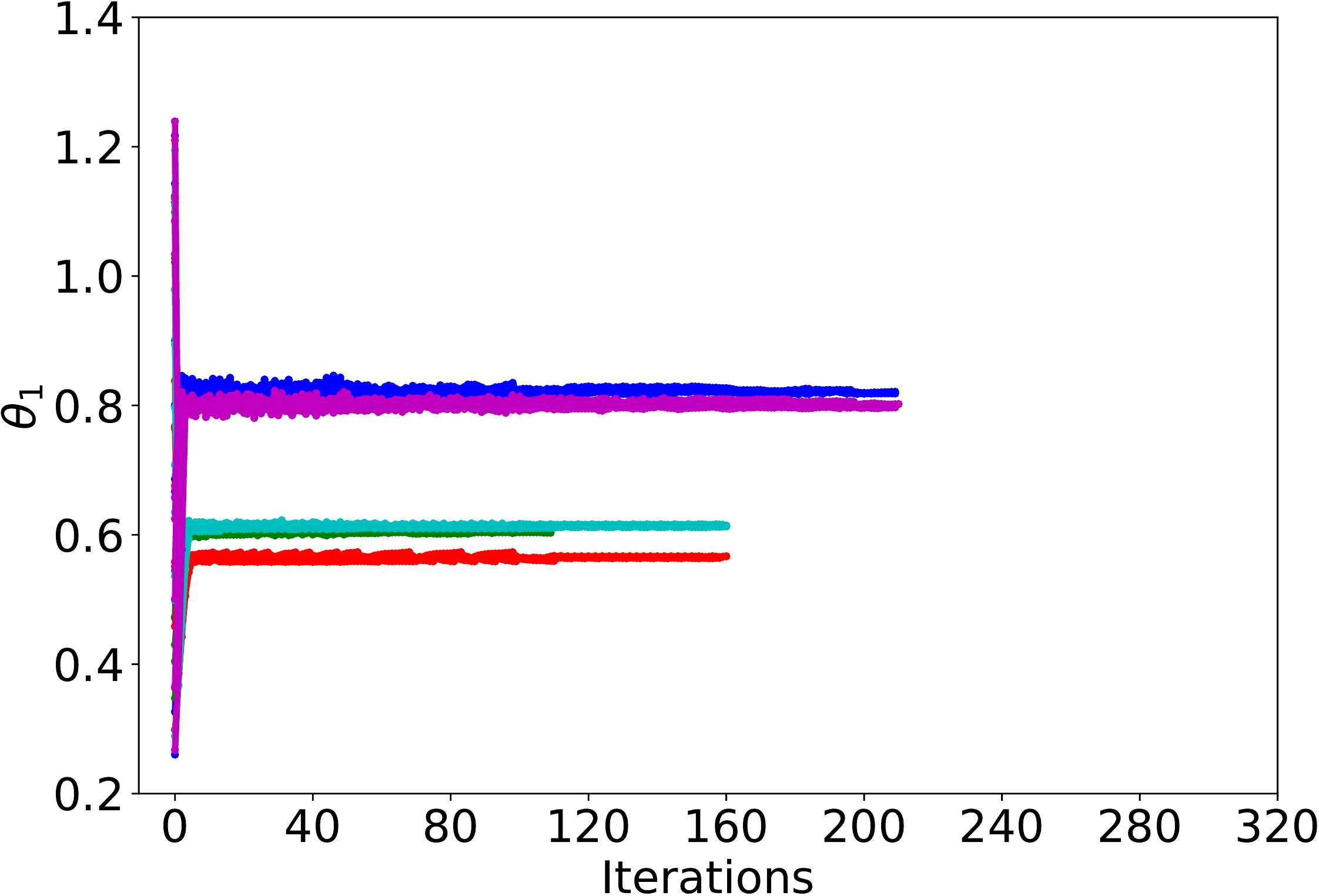}
\includegraphics[width=.425\textwidth]{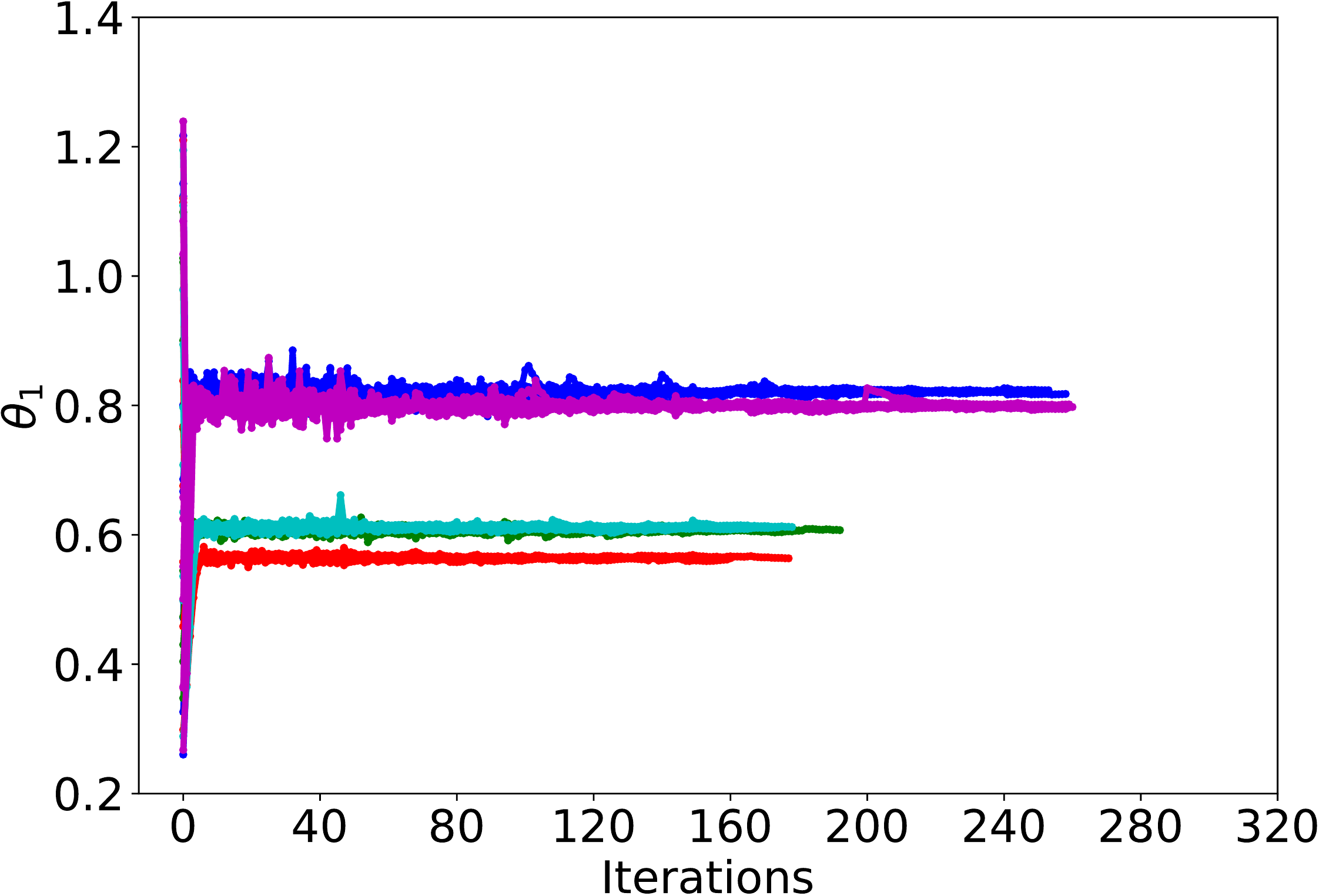}
}\qquad
\subfloat[\pmb{Geometric Brownian motion (GBM).}]{\label{fig:GBM_sgd}
\includegraphics[width=.425\textwidth]{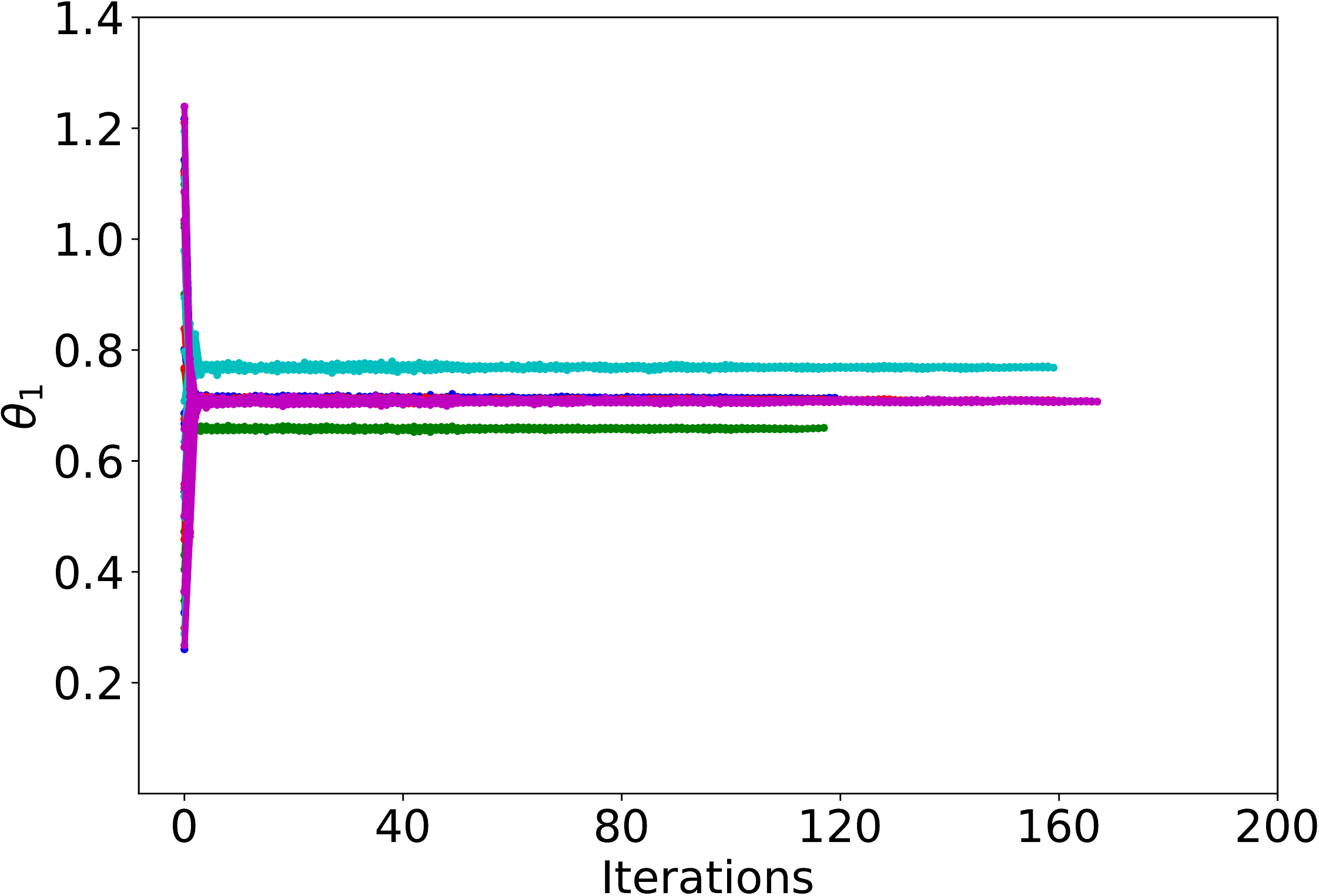}
\includegraphics[width=.425\textwidth]{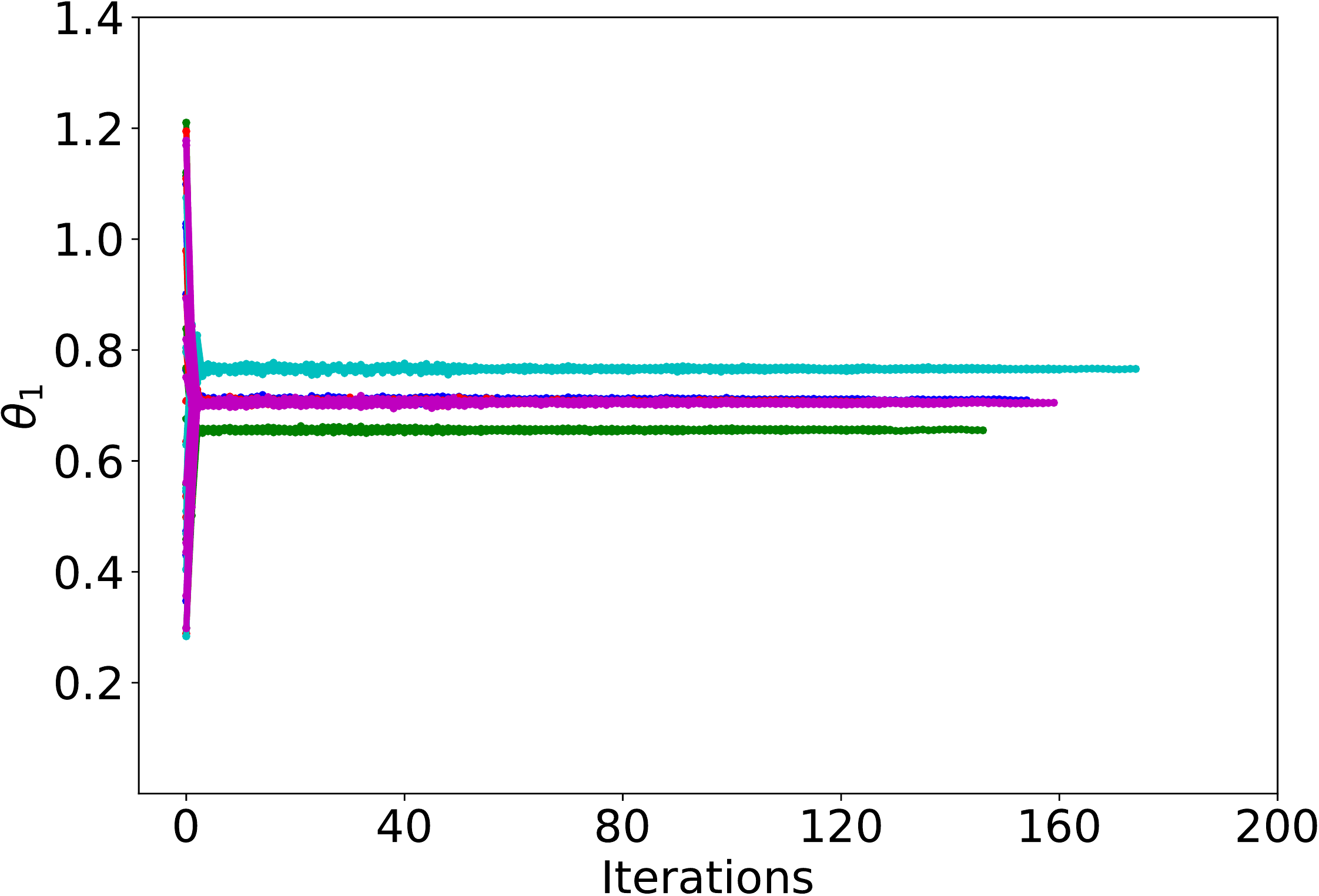}
}\qquad
\subfloat[\pmb{Lorenz model (LM)}.]{\label{fig:LM_sgd}
\includegraphics[width=.425\textwidth]{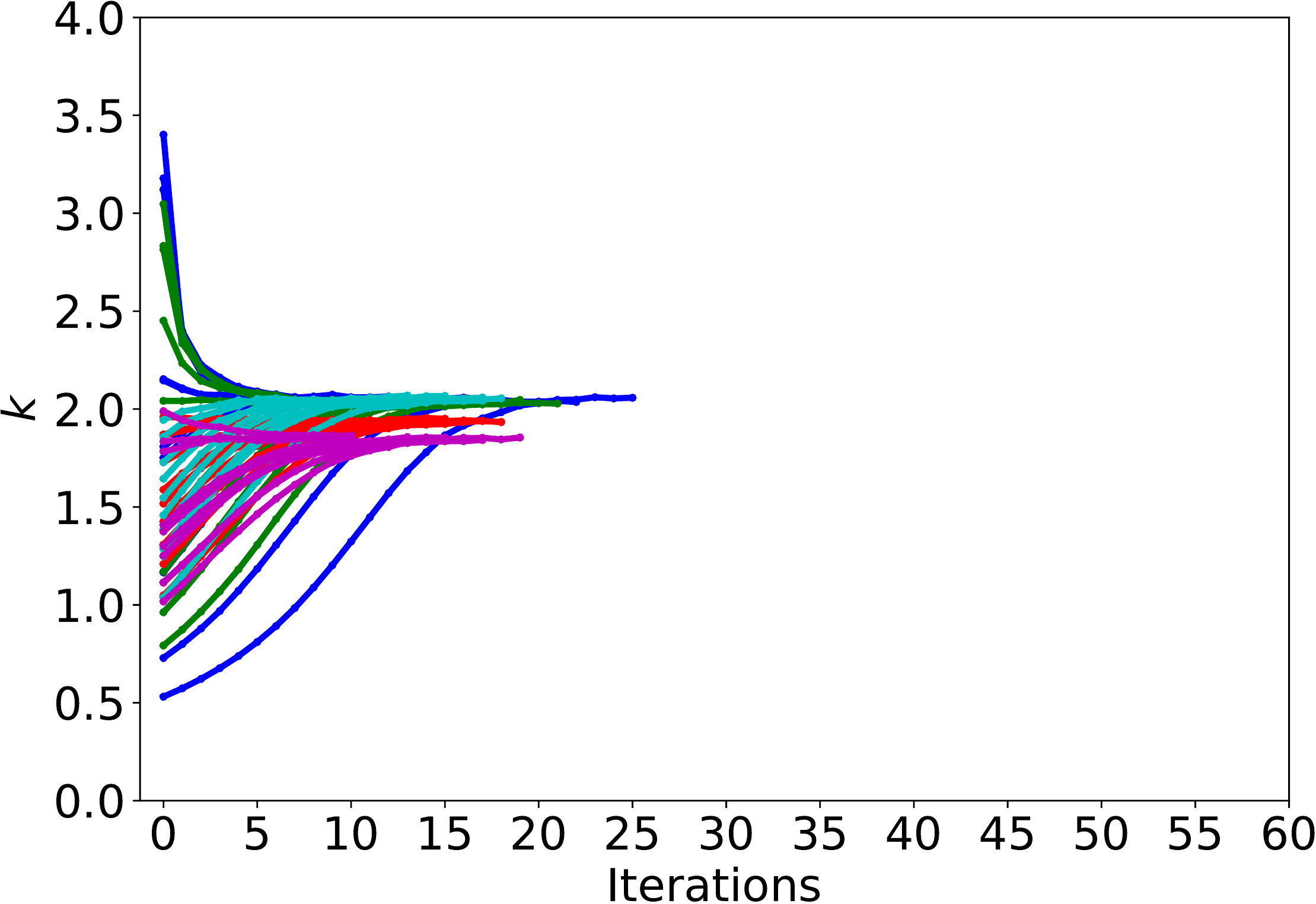}
\includegraphics[width=.425\textwidth]{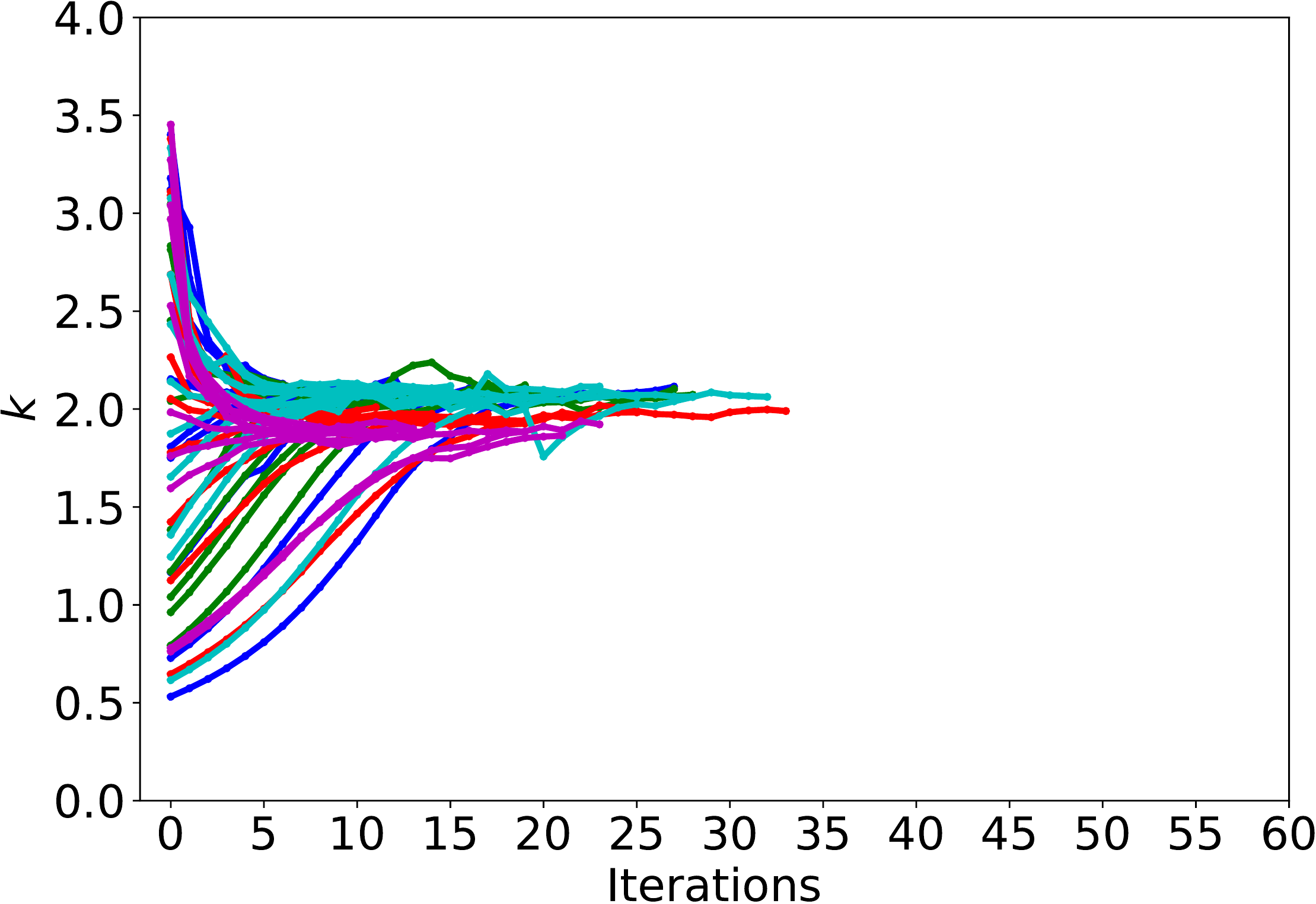}}%
\caption{\pmb{Stochastic gradient descent}. Given time series $Y^s_{t}$ with $s=\{1,\dots,5\}$, we compute $10$ repeats of Algorithm \ref{alg:SGD} with random initialization. \pmb{Left} \pmb{figure}: Algorithm \ref{alg:SGD} solved by Rhee-Glynn estimator \eqref{eq:psi_1}. \pmb{Right} \pmb{figure}: Algorithm \ref{alg:SGD} solved by unbiased estimator \eqref{eq:estimate_coup} with underlying empirical distribution.}
\label{fig:SGD}
\end{figure}

The unknown parameters estimated by SGD algorithm, $\theta_1$ for the OU and GBM cases and $k$ for the LM, are in Tables \ref{table:OU}, \ref{table:GBM}, and \ref{table:LM}. The inferred parameters are consistent with the model values. The iterations before meeting the stopping condition are higher for the unbiased estimator \eqref{eq:estimate_coup} with geometric distribution with respect to the other two estimators for its higher variance. Unbiased estimator \eqref{eq:estimate_coup} with empirical distribution and Rhee-Glynn estimator show a comparable number of iterations as displayed in \ref{fig:SGD} but unbiased estimator \eqref{eq:estimate_coup} with empirical distribution is more computationally expensive, coherently with previous analysis. 

\begin{table}
\centering
\begin{tabular}{|c | c c c |} 
 \hline
 Model & $\mu(\cdot)$ & $\alpha$ & $\beta$ \\ [0.5ex] 
 \hline\hline
 OU  & $\mathcal{U}_{[0.25,1.25]}$ & $5 \cdot 10^{-2}$ & $10^{-3}$  \\ 
 GMB & $\mathcal{U}_{[0.25,1.25]}$ & $2.5 \cdot 10^{-2}$ & $10^{-3}$  \\
 LM  & $\mathcal{U}_{[0.5,3.5]}$ & $1.5625\cdot 10^{-3}$ & $0.05$ \\ [1ex] 
 \hline
\end{tabular}
\caption{Parameters used in Algorithm \ref{alg:SGD} for each model.}\label{table:SGD}
\end{table}

\begin{table}
\centering
\subfloat[\pmb{Ornstein-Uhlenbeck (OU).}]{\label{table:OU}
\begin{tabular}{|c|ccc|ccc|ccc|}
\hline
$Y_t$  & \multicolumn{3}{c}{$\theta_1$} & \multicolumn{3}{|c|}{Iterations} & \multicolumn{3}{|c|}{Time [s]}\\
 \hline
 s  & Geom & Emp & RG  & Geom & Emp & RG & Geom & Emp & RG  \\
\hline
 1 & 0.82  & 0.82 & 0.82 & 220.3  & 222.3 & 169.5 & 2921 & 24573 & 267 \\
 2 & 0.60  & 0.60 & 0.60 & 122.5  & 166.2 & 60.9  & 1300 & 17245 & 86  \\
 3 & 0.56  & 0.56 & 0.56 & 87.1  & 151.6 & 74.3  & 1962 & 16541 & 100 \\
 4 & 0.61 & 0.61 & 0.61 & 140.9 & 167.5 & 87.9  & 2894 & 16752 & 120 \\
 5 & 0.80 & 0.80  & 0.80 & 196.1 & 240.2 & 185.2 & 495 & 27935 & 291 \\
\hline
\end{tabular}
}

\subfloat[\pmb{Geometric Brownian motion (GBM).}]{\label{table:GBM}
\begin{tabular}{|c|ccc|ccc|ccc|}
\hline
$Y_t$  & \multicolumn{3}{c}{$\theta_1$} & \multicolumn{3}{|c|}{Iterations} & \multicolumn{3}{|c|}{Time [s]}\\
\hline
 s  & Geom & Emp & RG & Geom & Emp & RG & Geom & Emp & RG \\
\hline
 1 & 0.71  &      0.71    & 0.71 & 150.7  &  124.2    & 108.5 & 1974 &  16011    & 212 \\
 2 & 0.65 &   0.65     & 0.65 & 136.1  &    115.4    & 105.4 & 2780 &  14794   & 206 \\
 3 & 0.70 &      0.70   & 0.70 & 164.6  &     112.4  & 119.7 & 1755 &     13302  & 234 \\
 4 & 0.76 &     0.76    & 0.76 & 172.     &     149.9  & 143.9 & 2274 &  19358   & 281 \\
 5 & 0.70 &     0.70    & 0.70 & 172.     &     142.6  & 141.4 & 2082 &   16642  & 275 \\
\hline
\end{tabular}
}

\subfloat[\pmb{Lorenz model (LM).}]{\label{table:LM}
\begin{tabular}{|c|ccc|ccc|ccc|}
\hline
$Y_t$  & \multicolumn{3}{c}{$\theta_1$} & \multicolumn{3}{|c|}{Iterations} & \multicolumn{3}{|c|}{Time [s]}\\
\hline
 s  & Geom & Emp & RG  & Geom & Emp & RG & Geom & Emp & RG  \\
\hline
 1 & 2.05     &    2.06   & 2.04 & 25.6 &    17.8     & 15.7 & 7983 &    52764     &  1191\\
 2 & 2.05     &    2.08   & 2.04 & 27.8 &   19.3     & 16.1 & 3958 &    56554       &  640\\
 3 & 1.95     &    1.96    & 1.93 & 22.3 &    18.5     & 14.2 & 6933 &    58394      &  1070\\
 4 & 2.06    &    2.07    & 2.05 & 27.7 &   21.3      & 15.6 & 5859 &    66665      &  690 \\
 5 & 1.85    &    1.88   & 1.85 & 18.7 &      16.1    & 15.  & 4675   &       49177      & 1133\\
\hline
\end{tabular}
}
\caption{Given time series $Y^s_{t}$ with $s=\{1,\dots,5\}$, we average over $10$ i.i.d. repeats of Algorithm \ref{alg:SGD} to estimate unknown parameter ($\theta_1$ for Ornstein-Uhlenbeck and Geometric Brownian motion cases, $k$ for Lorenz model), number of iterations before meeting stop criteria, and computation time in seconds.}
\end{table}


%

\subsubsection*{Acknowledgements}
The authors were supported by KAUST baseline funding.

\end{document}